# Devanagari transliteration: A complete character recognition and transliteration technique for Devanagari script

**Jasmine Kaur,**[a,*] **Vinay Kumar**[a]

[a]Department of Electronics and Communication Engineering, Thapar University, Patiala, Punjab, India.

**Abstract** Transliteration involves transformation of one script to another based on phonetic similarities between the characters of two distinctive scripts. In this paper, we present a novel technique for automatic transliteration of Devanagari script using character recognition. One of the first tasks performed to isolate the constituent characters is segmentation. Line segmentation methodology in this manuscript discusses the case of overlapping and skewed lines. Overlapping line segmentation is based on number of connected components which is made equivalent to number of individual lines in the image. Mathematical morphological operation, closing and dilation to be exact are used to limit skew angle variation range thereby expediting the projection profile method of skew correction. The presented skew correction method works for full range of angles. Character segmentation algorithm is designed to segment conjuncts and separate shadow characters. Presented shadow character segmentation scheme employs connected component method to isolate the character, keeping the constituent characters intact. Statistical features namely different order moments like area, variance, skewness and kurtosis along with structural features of characters are employed in two phase recognition process. After recognition, constituent Devanagari characters are mapped to corresponding roman alphabets in way that resulting roman alphabets have similar pronunciation to source characters.

**Keywords:** Image processing, segmentation, morphology, statistical and structural features.

*First Author, E-mail: tjasminekaur@gmail.com

## 1. Introduction

Devanagari script has its roots dated back thousands of years. Most of the Indian literature such as Bhagavad Gita, Vedas, Mahabharata, Ramayana is written in Devanagari. Such voluminous literature necessitates transliteration into roman script to make it more accessible to people who are unfamiliar with Devanagari. This paper attempts in automatic Romanization of Devanagari document using character recognition with the help of underlying statistical properties of alphabets. The statistical features include area, mean, variance, kurtosis and skewness together with structural features of alphabets.

*1.1. Devanagari word composition*

Devanagari word comprises of header line, character set and modifiers as displayed in the Fig. below:

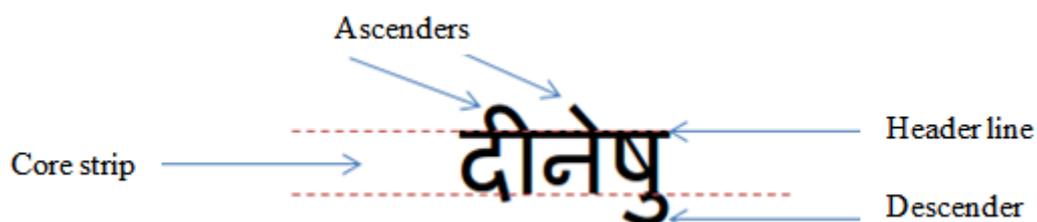

**Fig. 1** Character zones in Devanagari script

A detailed description about Devanagari font structure can be found at URL in Ref. 1.

## 2. Related work

After extensive research in recognition of roman scripts, a fair amount of work has been reported in recognition of Devanagari script.

The pioneers being R.M.K Sinha and V. Bansal[2-4] who presented various aspects of Devanagari character recognition (CR). Over the course of six years the duo presented techniques on how to partition touching and fused Devanagari characters, use the knowledge of Devanagari character shapes for recognition and how to integrate structural and statistical knowledge about characters in CR system. The presented CR system utilizes height and width information along with collapsed horizontal projections of character for segmentation and recognition. In Ref. 3 R.M.K Sinha et al. devised a complete Devanagari CR system which used horizontal projection of the character for segmentation. This technique made the use of horizontal zero crossing, number and position of end points and structural strokes of a character as feature extraction method. In the case of horizontal zero crossings, the number of transitions from black to white pixels are calculated. The character image is divided in 3x3 blocks to note the vertex points. In Ref. 2 the duo designed a Devanagari CR system for varying fonts and size. Some of the statistical knowledge sources are acquired a priori by an automated training process while others are extracted from the text during processing. Later they proposed a segmentation method[4] for touching and fused Devanagari characters. This method uses a two pass algorithm to segment the composite characters. First, the statistical information about width and height of a character is used to determine whether the character box consists of composite characters or not. In second pass, these composite characters are further segmented using horizontal and vertical projection profiles. U. Garain et al.[5] developed an algorithm for effectively selecting a segmentation column in touching characters. This method uses fuzzy multifactorial analysis to isolate the characters which require further segmentation.

Segmentation is critical step in recognition system. Several segmentation methods have been tried and tested by researchers. In a line segmentation technique proposed by U. Pal et al.[6] hill and valley points (crest and troughs respectively) are found in a smoothed projection profile. Smoothed horizontal profile is obtained by changing white run length that is less than a threshold into black. A new methodology for line segmentation is given by U. Pal and S. Dutta[7]. In this case text image is divided into vertical stripes and for each strip potential piece-wise separating line is found which is utilized to segment to the lines. In another line segmentation algorithm[8] header line detection along with base line detection is employed. S. Kompalli et al.[9] explained several challenges faced in Devanagari optical character recognition (OCR) like fused ascender segmentation, descender segmentation due to varying character height and corresponding modifier placement and problems faced during core component separation. Rather than using rule-based segmentation S. Kompalli et al.[10] employed recognition driven segmentation. This method generates block adjacency graph for a particular character by dividing it into horizontal runs. Segmentation is performed by selecting subgraphs from the block adjacency graph representation of word or character images. S. Pal et al.[11] in 2010 presented a technique to segment overlapping lines by tracing contour points of smeared text lines.

Feature extraction is process of calculating appropriate measures to characterize character image. Devanagari OCR researchers have utilized very rich set of features, incorporating geometric moment invariants, projection histograms, geometric or structural features etc. U. Pal et al.[12] calculated histogram for directional feature obtained from each segmented contour image blocks. In another method T. Wakabayashi et al.[13] enhanced the accuracy of technique used in Ref. 12 by weighing the features by F-ratio. In Ref. 14 researchers used intersection and open end points for recognition purpose. S. Arora et al.[15] utilized shape and contour information to chain code histogram, shadow and view based features. For shadow features, shadow of character segment is computed on perpendicular sides of octant. View based feature is a set of points that plot one of four projections of character image. R. J. Ramteke et al.[16] employed moment invariants and 2x2 diagonal zoning as feature extraction methods. The recognition scheme by M. Hanmandlu et al.[17] utilizes exponential membership function which represents features consisting normalized distance obtained by dividing the character image into 6x4 grid. Reference 18 gives detailed survey of Devanagari script recognition techniques.

A. Papandreou et al.[19] in 2011 proposed a hybrid technique that uses both vertical projection and bounding box technique to determine the skew in a document with latin characters. For a non-skewed document great peaks are observed in the certain columns and the area of bounding box is minimum. The ratio of summation of black pixels in a column to the area of bounding box is maximised, by rotating the image, to determine the skew angle. Shutao Li et al.[20] in 2007 presented a method for skew detection using wavelet decomposition and projection profile method. This scheme works better if we know the layout of the document beforehand. B. V. Dhandra et al.[21] in 2006 presented a method which uses morphological technique dilation to fill the space between the characters using a fixed appropriate structuring element. The dilated text lines are then region labelled to find orientation angle for different region.

**3. Proposed methodology and challenges faced at various stages.**

The present manuscript focuses on automatic transliteration of Devanagari script. We consider the script as image generated with the help of scanner or photographed. Next sections explain various stages of proposed Devanagari recognition scheme. During document processing several possible challenges and errors are perceived. The possible errors for different phase have been indicated in Fig. 2. The present work discusses each of these challenges and possible ways to overcome them while developing automatic transliteration system.

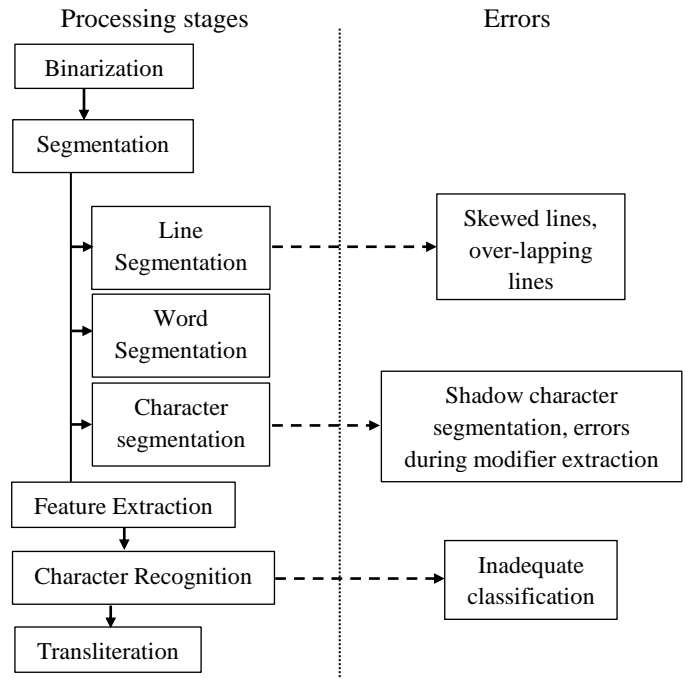

**Fig. 2** Stages of Devanagari recognition process and errors in different stages.

Transliteration requires documents to be segmented into characters, feature extraction and recognition of these characters followed by successive Romanization of Devangari characters and thus needs to perform following tasks:

1. Binarization of Devanagari document.
2. Extracting individual lines from the document image.
    2.1. Segmentation of un-skewed lines.
    2.2. Segmentation of overlapping lines-*(Sec. 5.2.2).*
    2.3. Segmentation of skewed lines-*(Sec. 5.3.1.)*
3. Extracting word subimages using vertical projection.
4. Preliminary character segmentation using projection profile method.
5. Segmentation of lower modifier-*(Sec. 7.1.1.)*
6. Segmentation of conjuncts-*(Sec. 7.2.1.)*

7. Segmentation of shadow characters- *(Sec. 7.3.)*

8. Feature extraction and classification of modifiers-*(Sec. 8.)*

9. Feature extraction for character recognition-*(Sec. 9.)*

10. Recognition of extracted characters-*(Sec. 10.)*

11. Transliteration of recognized characters- *(Sec. 11.)*

**4. Binarization**

A image of Devanagari script $I_G$ is converted into an inverted binary image I'$_{HxW}$ by using fundamental global thresholding technique[23]. Binary image is inverted so that relevant character information is represented by value '1' whereas redundant background information is represented by '0'. The histogram of grayscale image, with intensity values $I_G(m,n) \epsilon [0,1…255]$, for intensity value *i* is found using Eq. (1).

$$H(i) = \sum \delta(I_G(m,n) - i) \qquad (1)$$

Since image consists of text over plain black background, we do not consider background with patterns, resulting histogram will be concentrated largely at the extreme ends of grayscale image and difference Δ (given by Eq. (2))

$$\Delta = \max_{i\epsilon[0,127]} H(i) - \max_{i\epsilon[128,255]} H(i) \qquad (2)$$

is quite large.

Through experimentation it is found that suitable value of threshold *T* lies in the vicinity of $\Delta/2$ and different threshold values in this effective threshold range have little to no effect on the resulting image I'$_{HxW}$ obtained using Eq. (3).

$$\text{I'}_{HxW} = \sim\text{I}_{HxW}(m,n) = \begin{cases} 1 & if\ I_G(m,n) \leq T \\ 0 & if\ I_G(m,n) > T \end{cases} \qquad (3)$$

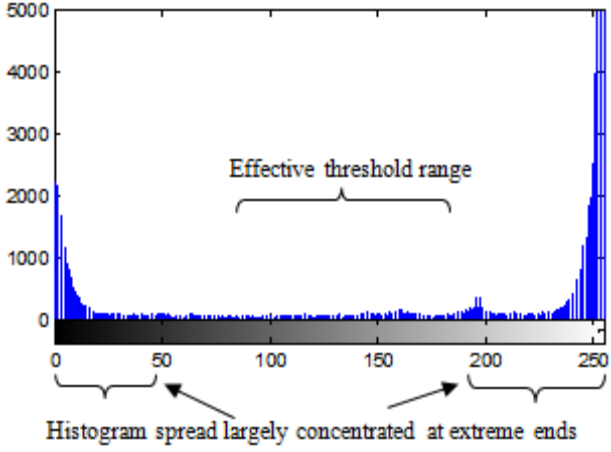
(a)

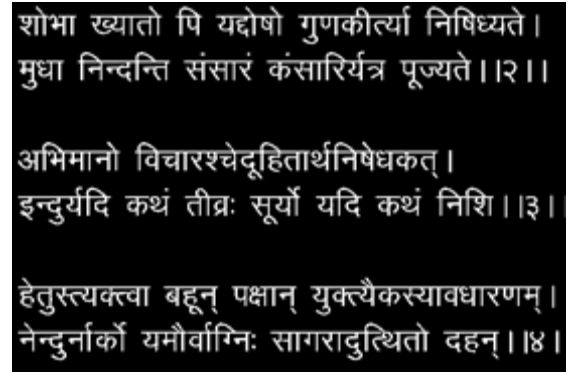
(b)

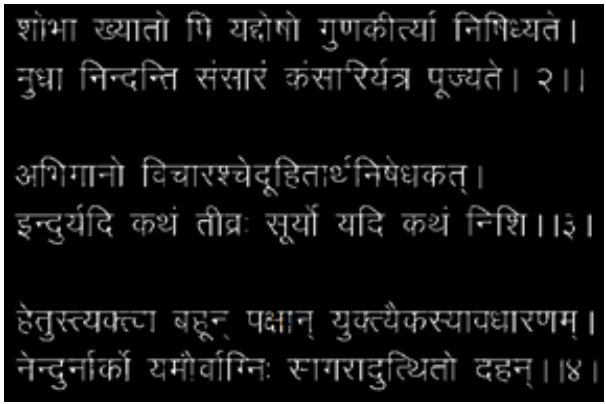
(c)

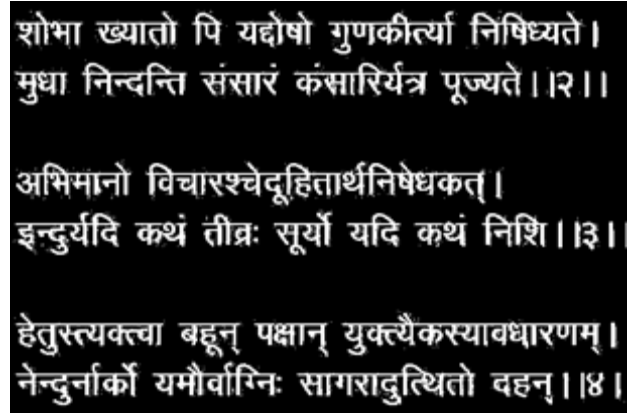
(d)

**Fig. 3** (a) Histogram of input image (b) output binary image with threshold $T$=170 (c) output binary image with threshold $T$=30 (d) output binary image with threshold $T$=220.

Image $I'_{HxW}$, white text on black background, is used in all of the following subsequent steps. Figure 3 shows result of binarization on $I_G$. When using a threshold value lying in a threshold region, precise binarized image is obtained with no broken or smeared characters, Fig. 3(b). However, choosing a threshold value lying beyond threshold range gives an image where background is not distinctively separated from foreground, Fig. 3(c), whereas a threshold value lying to the left of threshold range gives fragmented characters, Fig. 3(d).

## 5. Line segmentation

*5.1. Segmentation of unskewed lines*

Decomposition of image $I'_{HxW}$ into individual lines is carried out by computing horizontal projection $HP(m)$ using Eq. (4), if the document has skew we first de-skew (refer Sec. 5.3.1) it to remove the tilt.

Horizontal projection is a histogram calculating number of white pixels in each row of an image. Rows corresponding to zero *HP(m)* value are used to isolate the adjacent lines shown in Fig. 4.

$$HP(m) = \sum_{n=1}^{W} I'(m,n) \ \forall \ m \in [1, H] \tag{4}$$

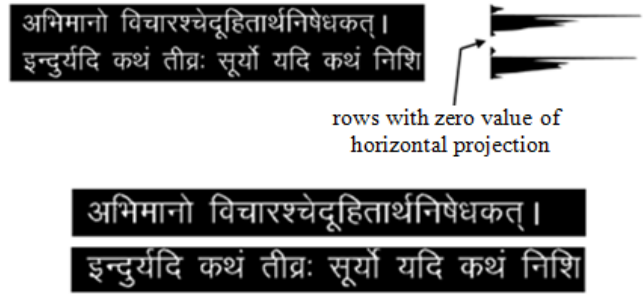

**Fig. 4** Line segmentation (a) Input image $I'_{HxW}$ with its horizontal projection (b) Constituent segmented lines.

*5.2. Segmentation of overlapping lines*

In Devanagari when descenders of upper line overlap with ascenders of lower line, two lines partially overlap. It is not possible to separate them merely by drawing a horizontal line due to the absence of segmenting row in horizontal projection as shown in Fig 5(a). Instead of segmenting lines by drawing a parallel cut (which leaves behind fragmented modifiers of another line as shown in Fig 5(b)) or by making a contour[14] we use the concept of connected component (CC) to separate these lines.

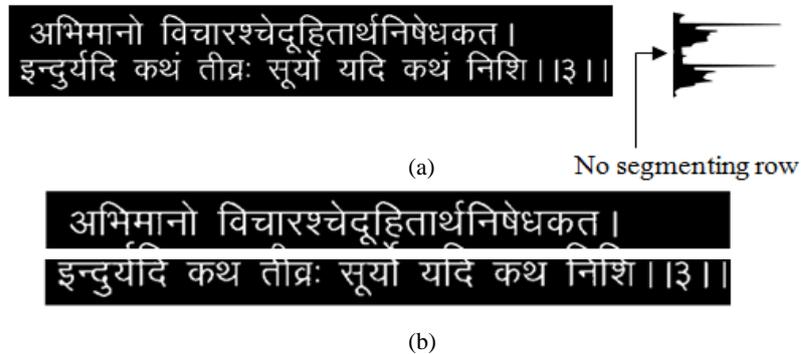

**Fig. 5** (a) Overlapping lines with horizontal projection (b) segmented lines by drawing a horizontal parallel cut

*5.2.1. Connected component labelling*

To isolate the overlapping lines, we need to determine CCs in such a way that number of CCs is *analogous to number of lines* in a document image. CCs labelling sweeps the entire document image and groups the foreground (white) pixels into components based on pixel connectivity, *i.e.* all pixels in a connected component share same pixel intensity value and are linked with each other. When finding number of connected elements in an image generally two types of pixel connectivities[22] are supported, 4-pixel connectivity shown in Fig. 6 (a) and 8-pixel connectivity shown in Fig. 6 (b).

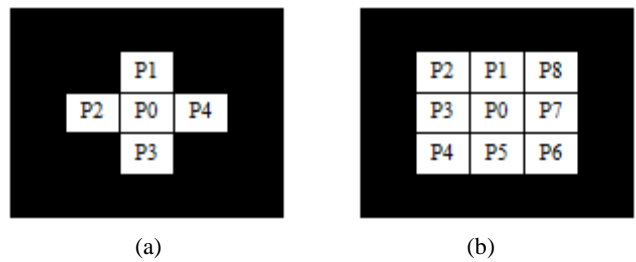

(a)             (b)

**Fig. 6** (a) Example of 4-pixel connectivity where pixels P0, P1, P2, P3 and P4 form a 4-pixel connected component (b) One connected component based on 8-pixel connectivity

*5.2.2. Overlapping lines segmentation methodology*

Using 8-pixel connectivity overlapping lines in the image $I'_{ncc}$ can be isolated using following steps:

1. Find the number of 8-connected elements[22] $N_{cc}$ and corresponding pixels $I^i_{cc}$ associated with each component $n^i_{cc}$, where $i$=1, 2…….$N_{cc}$.

2. Image $I_{ncc}$ containing overlapping lines with detached modifiers removed can be found using Eq. (5) and is shown in Fig. 8(b).

$$I_{ncc} = (I_{cc}^i = 0 \Leftrightarrow card(I_{cc}^i) < 10) \tag{5}$$

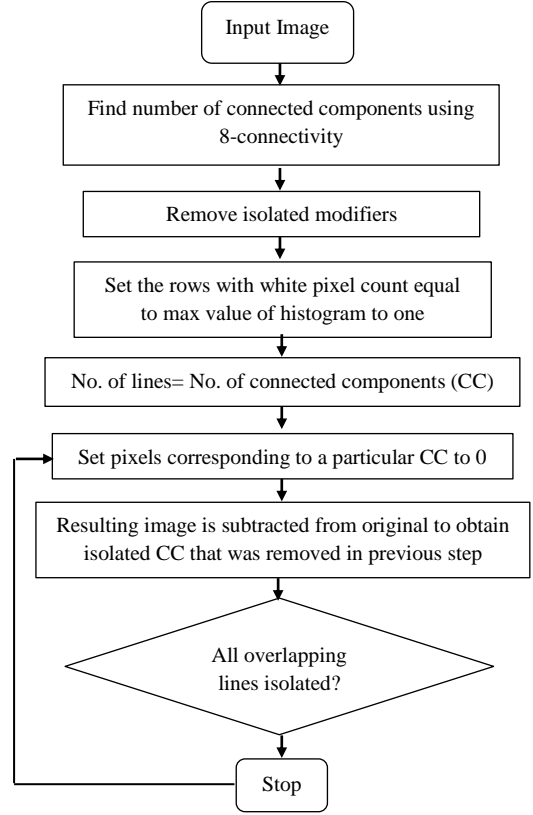

**Fig. 7** Flowchart representing method to isolate overlapping line

3. For image $I_{ncc}$, horizontal projection histogram is found using Eq. (6), (7) and Eq. (8). It results in grouping of all the words, in a particular line, into one wholesome 8-connected component. After this operation number of CC correspond to number of individual lines in an image as shown in Fig 8(c).

$$HP(m) = \sum_{n=1}^{W} I_{ncc}(m,n) \ \forall \ m \in [1,H] \tag{6}$$

$$Rows = (m|HP(m) = HP_{max} \ \forall \ m) \tag{7}$$

where $HP_{max}$ denotes the maximum value of horizontal projection $HP(m)$.

$$I_{ncc}(m,n) = 1 \ \forall \ n,m \in Rows \tag{8}$$

4. Image $I_{ncc}$, obtained from Step 3, pixels corresponding to a specific CC, which needs to be isolated, are set to zero. The resulting image is obtained using Eq. (9) and is shown in Fig 8(d).

$$I_{ncc}(I_{cc}^i) = 0; \ where \ i \in [1, N_{cc}] \tag{9}$$

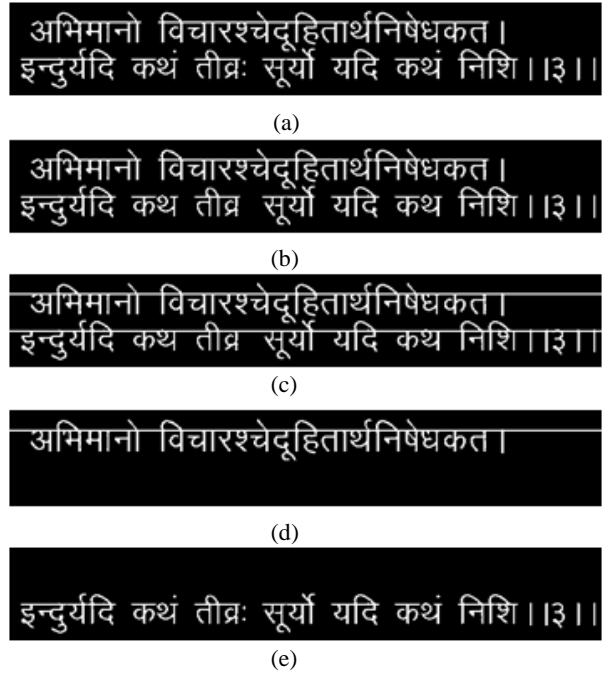

**Fig. 8** (a) Overlapping lines to be segmented (b) isolated modifiers removed (c) corresponding words of a line grouped into one connected component (d) image after removing one connected component (e) segmented line

5. Subtracting image $I_{ncc}$, obtained from step 4, from original image $I'_{ncc}$ results in an image containing isolated line $(n_{cc}^i)$ which was removed in Step 4 as shown in Fig 8(e).

$$n_{cc}^i = I'_{ncc} - I_{ncc} \tag{10}$$

6. Step 4 and 5 are repeated until all the overlapping lines are segmented.

*5.3. Segmentation of skewed lines*

For segmentation of skewed lines, alignment of text lines with respect to horizontal axis is first corrected and then horizontal projection profile method, Sec. 5.1, is used to isolate the lines.

Traditional projection profile methods are extremely slow and tend to fail for larger skew angles as the input document has to be rotated iteratively through fairly large range of angles. Projection profile methods are well suited to estimate skew angles within $\pm\ 10^0$. In the proposed improved approach of skew correction, time to compute skew angle is substantially reduced by using morphologically operations to limit the effective range of resulting skew angle variation of a document to $\leq 15^0$. So the document needs to be rotated by an angle in this particular range, thereby increasing the computational speed. Along with increasing the speed and accuracy, operation limit for our proposed method is improved to full range of skew angle detection, i.e. $-180^0$ to $180^0$.

The effect of morphological operations[23] used in proposed algorithm is explained below:

**Dilation:** Linear structuring element (SE) symmetric with respect to centre and half the width of input image is chosen in the proposed scheme.

The initial angle of the line is $10^0$ and direction of orientation of SE depends on whether the image is skewed in clockwise or anti clockwise direction. The alignment of the line SE is increased iteratively by a constant value during subsequent steps of proposed algorithm.

Observing Fig. 9 we conclude that changing the orientation of line SE with respect to the text lines changes the number of bounding box in an image. Through experimentation it is seen that when angle between a text line and line SE is a small value then the number of bounding box is approximately equal to number of lines in an image. As we keep on increasing this angle difference,

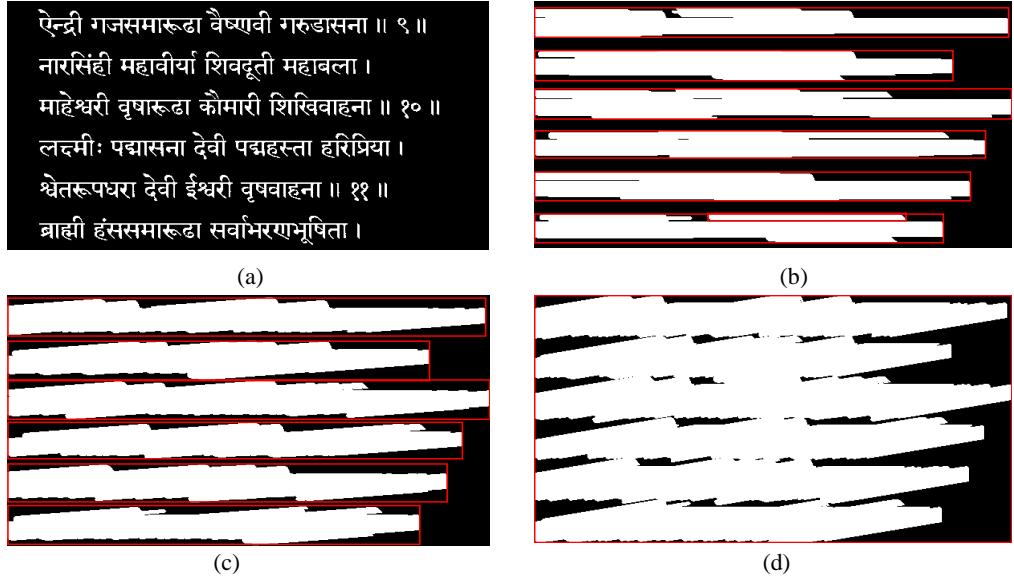

**Fig. 9** Example of dilation with line SE of varying angles (a) original image (b) dilation with SE at angle of $0^0$ (c) dilation with SE at angle of $5^0$ (d) dilation with SE at angle of $10^0$

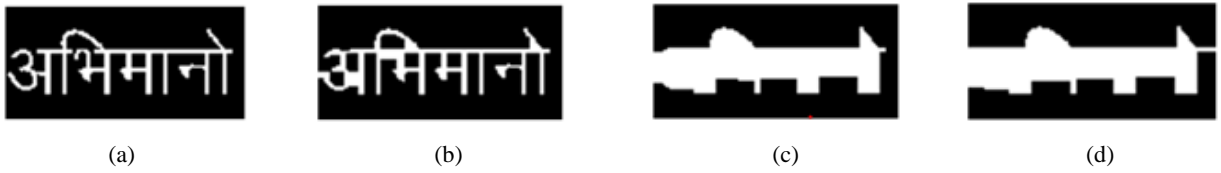

**Fig. 10** Example of closing with SEs of different size (a) original image (b) closing with SE of size 5x5 (c) closing with SE of size 10x10 (d) closing with SE of size 15x15

number of Bounding box (BB) decrease in number until finally becoming equal to 1 above a certain constant angle difference.

**Closing:** Performing closing operation on an image fills the background regions between the characters causing them to smear into solid white blobs. Performing closing operation on an image fills the background regions between the characters causing them to smear into solid white blobs. As seen in Fig 10(b), if the size of SE is too small then the words are not smeared perfectly. So, the size of SE must be selected in such a way that it is large enough to smear the words into perfect solid blobs.

*5.3.1. Skew correction methodology*

The proposed algorithm for de-skewing image $I_{skew}$ using morphological operations, depicted in flowchart of Fig. 11, is given in steps below:

1. Close the image $I_{skew}$, using Eq. (11), with square structuring element $S_{sq}$ to form solid text blobs corresponding to each text word.

$$I_{close} = I_{skew} \bullet S_{sq} = (I_{skew} \oplus S_{sq}) \ominus S_{sq} \tag{11}$$

where $\oplus$ and $\ominus$ denote dilation and erosion and can be found using Eq. (12) and (13).

$$I_{dilate} = I_{skew} \oplus S_{sq} = \{i + s | i \in I_{skew} \land s \in S_{sq}\} \tag{12}$$

$$I_{erode} = I_{skew} \ominus S_{sq} = \{i | \forall s \in S_{sq}, i + s \in I_{skew}\} \tag{13}$$

2. Compute horizontal projection profile of closed image and its peak value $HP_{max}^{org}$, using Eq. (14). Closed image and its corresponding projection profile are shown in Fig 12(c) and Fig. 12(d), respectively. Comparing horizontal projection of original image in Fig 12(b) with closed image projection in Fig. 12(d) we notice that peaks are more prominent in case of closed image projection.

$$HP_{max}^{org} = \max\left(\sum_{n=1}^{W} I_{close}(m,n) \ \forall \ m \in [1,H]\right) \tag{14}$$

3. Rotate the image $I_{close}$ in clockwise and anticlockwise direction by a small angle, e.g. $2^0$. Respective rotated images $I_{close}^C$ and $I_{close}^A$ are shown in Fig. 12(e) and Fig. 12(g).

4. Find horizontal projection profile of rotated images and their corresponding maxima $HP_{max}^{anti}$ and $HP_{max}^{clock}$ using Eq. (15) and (16) respectively. If peak of anticlockwise rotated image is greater than clockwise rotated image, i.e., $HP_{max}^{anti} > HP_{max}^{clock}$, image is skewed in clockwise direction and needs to be rotated in anticlockwise direction by a certain angle to remove the tilt and vice versa. For example, peak value $HP_{max}^{anti}$ of projection in Fig. 12(h) is less than $HP_{max}^{clock}$ value of Fig. 12(f), the image is skewed in counter clockwise direction.

$$HP_{max}^{clock} = \max\left(\sum_{n=1}^{W} I_{close}^C(m,n) \ \forall \ m \in [1,H]\right) \tag{15}$$

$$HP_{max}^{anti} = \max\left(\sum_{n=1}^{W} I_{close}^{A}(m,n) \ \forall \ m \ \in [1,H]\right) \tag{16}$$

$$\theta_{skew} < 0 \iff HP_{max}^{anti} > HP_{max}^{clock} \tag{17}$$

5. For anticlockwise skewed image dilation, using Eq. (18), is performed using a line structuring element $S_{line}$ angled at $\alpha°$ as shown in Fig. 12(i). For a clockwise skewed image dilation is performed using a line SE angled at $-\alpha^0$. We have chosen $\alpha = 10°$ because it's simpler to increment angle as a multiple of 10.

$$I_{dilate} = I_{skew} \oplus S_{line} = \{i + s | i \in I_{skew} \wedge s \in S_{line}\} \tag{18}$$

6. Number of enclosing BB is determined for resulting dilated image. For an image shown in Fig. 12(i) BB count is equal to 1.

7. If number of BB in *Step 6.* is greater than one, it implies that line SE and text lines of an image are oriented approximately at same angle and lower limit of skew angle variation range is assigned a value of $1^0$. Upper limit in this case is found by further dilating the skewed image using Eq. (19). The line SE element we use is of same length as in previous step and its orientation angle is iteratively increased by $\beta$ until encompassing BB number equals 1. SE orientation angle, for which BB enclosing the dilated image is equivalent to 1, forms upper limit of skew angle variation range. We have chosen $\beta = 10°$ because it is easier to increment angle as a multiple of 10 and is also time efficient. Choosing an angle less than 10° increases number of iteration and hence increases time consumed. Choosing an angle more than 10° increases effective skew angle variation range, thereby increasing time consumed.

8. When BB count in *Step 6.* is equivalent to one, orientation angle difference between line SE and text lines is greater than $\beta$. To find lower limit angle skewed image is dilated, using Eq. (18), with SE as used in *Step 6*, and its angle is incremented by $\beta$ for each dilation until number of BB is greater than one. Alignment angle of SE for which dilated image is enclosed by a single BB is taken as lower limit and image is continued to be dilated iteratively until enclosing BB number becomes 1 again. This final SE orientation angle is considered as upper limit of skew angle variation range. Limiting the skew angle variation to a much smaller range expedites the whole process of finding document tilt. For example, observing images in Fig. 12(j) through Fig. 12(l), range R is found to be $[40^0:55^0]$.

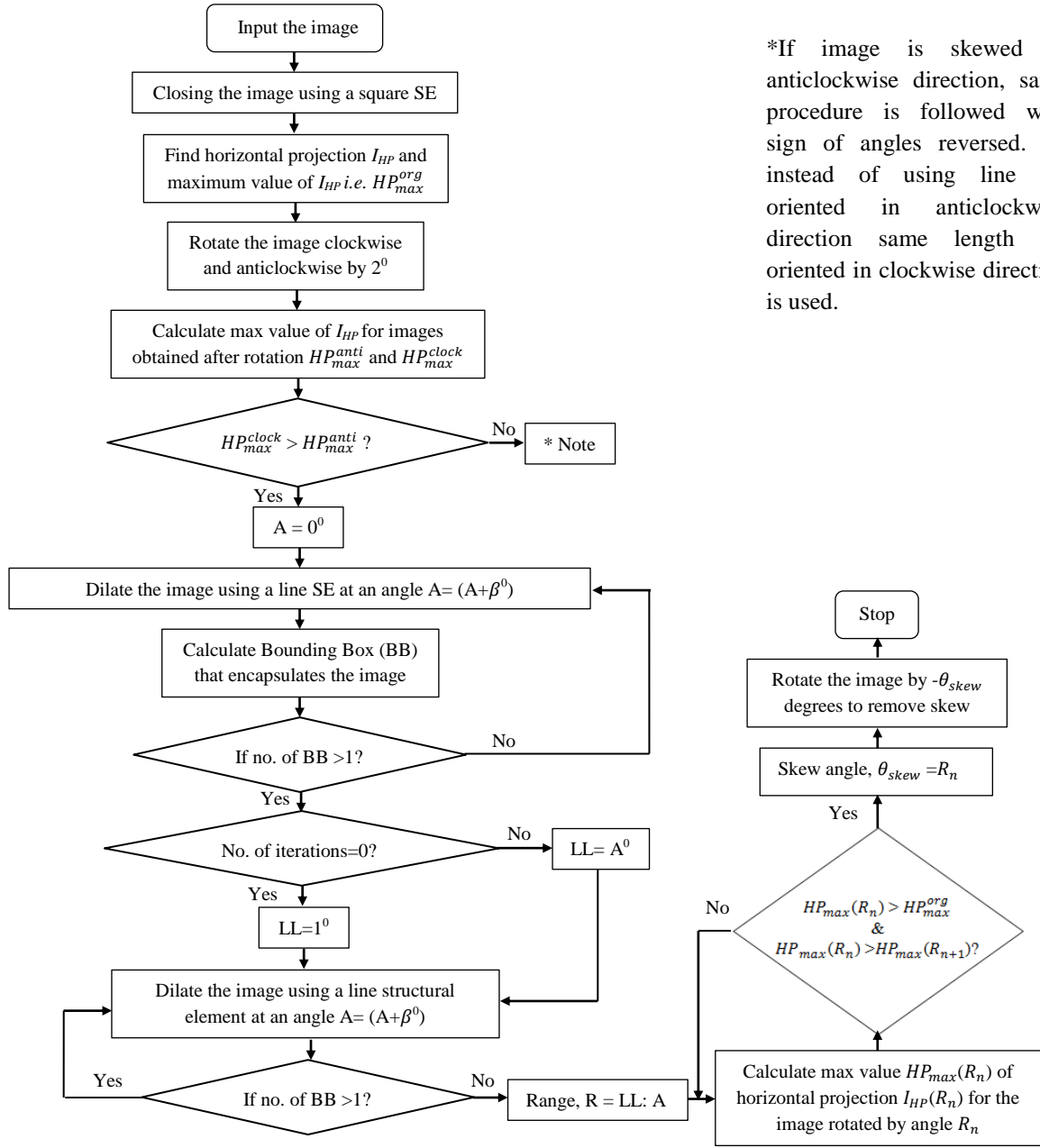

**Fig. 11** Flowchart depicting skew correction methodology

9. Once the final range of skew angles ($R$) is computed, closed image $I_{close}$, obtained in step 2, is rotated iteratively through this range and horizontal projection profile peak $HP_{max}(R_n)$ is calculated for angle $R_n$ until $HP_{max}(R_n)$ is both greater than $HP_{max}^{org}$ found in Step 2, and peak value $HP_{max}(R_{n+1})$ found for next angle in the range. $R_n$ represents $n^{th}$ angle value in range R.

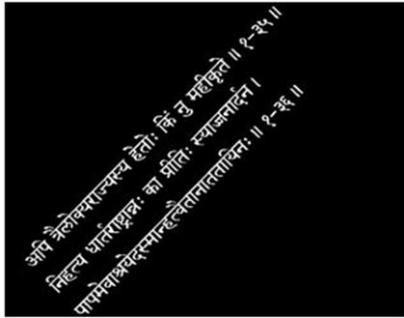
(a)

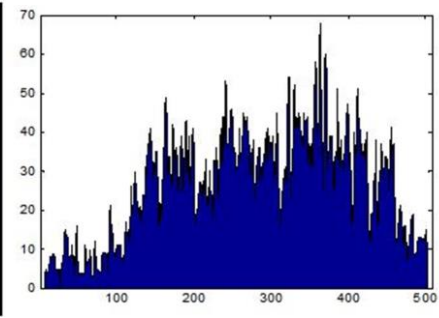
(b)

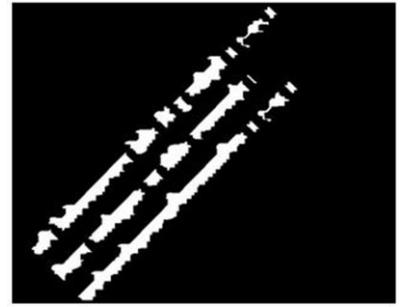
(c)

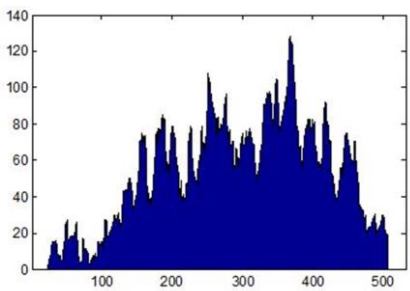
(d)

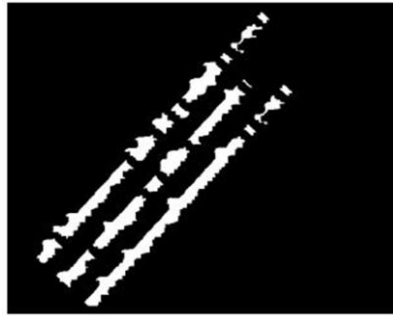
(e)

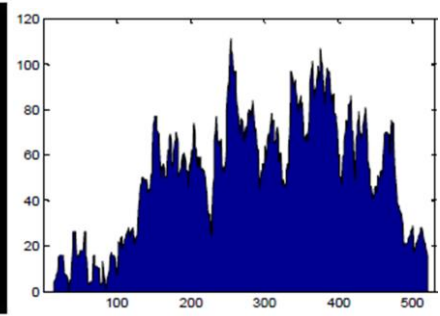
(f)

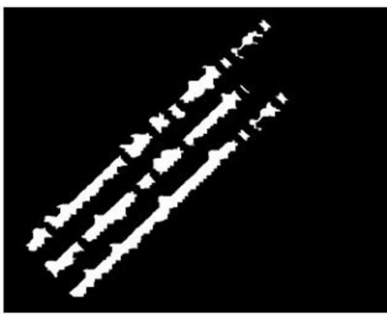
(g)

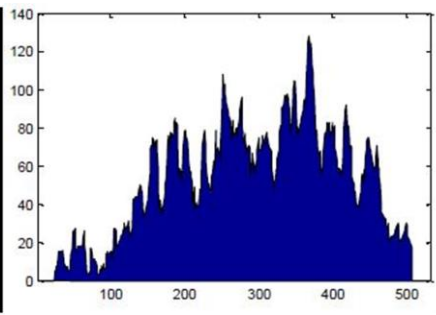
(h)

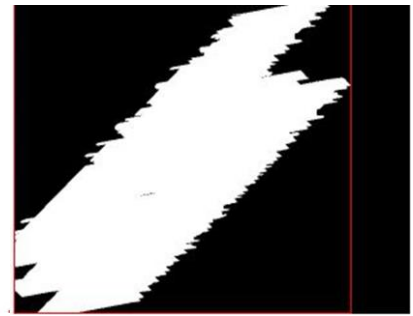
(i)

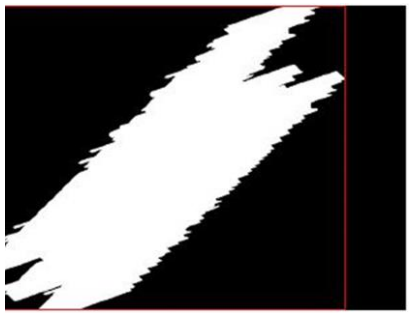
(j)

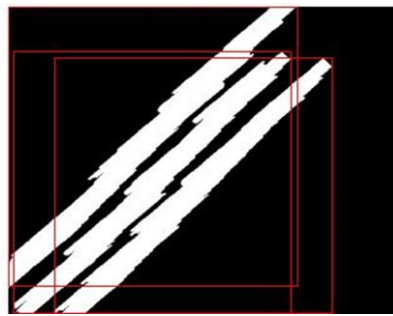
(k)

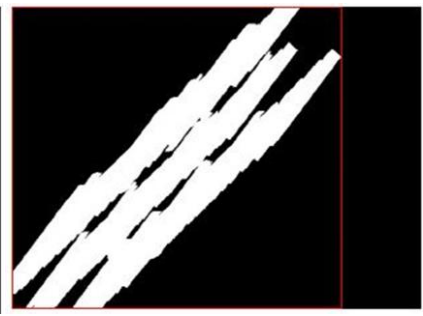
(l)

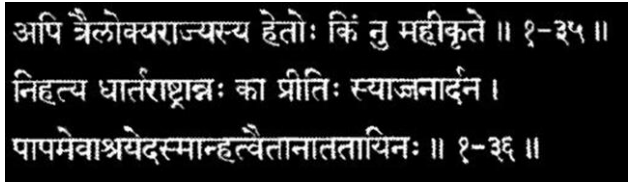

(m)

**Fig. 12** Skew correction of anticlockwise aligned image (a) original image (b) horizontal projection of original image (c) closed image using square SE of size 15x15 (d) horizontal projection of skewed image (e) image rotated clockwise by $2^0$ (f) horizontal projection of clockwise rotated image with peak value 128 (g) image rotated anticlockwise by $2^0$ (h) horizontal projection of anticlockwise rotated image with peak value 111 (i) image dilated using line SE of size 100 at angle $10^0$ (j) image dilated using line SE of size 100 at angle $20^0$ (k) image dilated using line SE of size 100 at angle $40^0$ (l) image dilated using line SE of size 100 at angle $55^0$ (m) image corrected by rotating original image by $-46^0$.

Angle $R_n$ satisfying Eq. (20) is resulting anticlockwise tilt of document image with respect to horizontal direction and this angle can be corrected by rotating the image by an angle of $-\theta_{skew}$. Final unskewed image, $I_{us}$ is given in Fig. 12(m).

$$HP_{max}(R_n) = \max\left(\sum_{n=1}^{W} I_{close}(m,n) \ \forall \ m \ \in [1,H]\right) \qquad (19)$$

$$\theta_{skew} = R_n \Leftrightarrow \left(HP_{max}(R_n) > HP_{max}^{org} \wedge HP_{max}(R_n) > HP_{max}(R_n+1)\right) \qquad (20)$$

For image skewed in clockwise direction Steps 5 through 9 are followed in order, the only difference is that line SE orientation angles by which image is dilated have opposite sign i.e., image is dilated using SE oriented in clockwise direction. Once the image skew is corrected, next part involves determining whether image $I_{us}$ has been flipped upside down or not.

Flip identification:

Flowchart in Fig. 13 indicates necessary steps for flip identification and are explained below.

1. Segment a single text line, say $L_{us}$ using method mentioned in *Sec. 5.1.* and crop it to its minimum bounding box dimensions.

2. Find horizontal projection histogram, $HP_L(m)$, for image $L_{us}$ using Eq. (4).

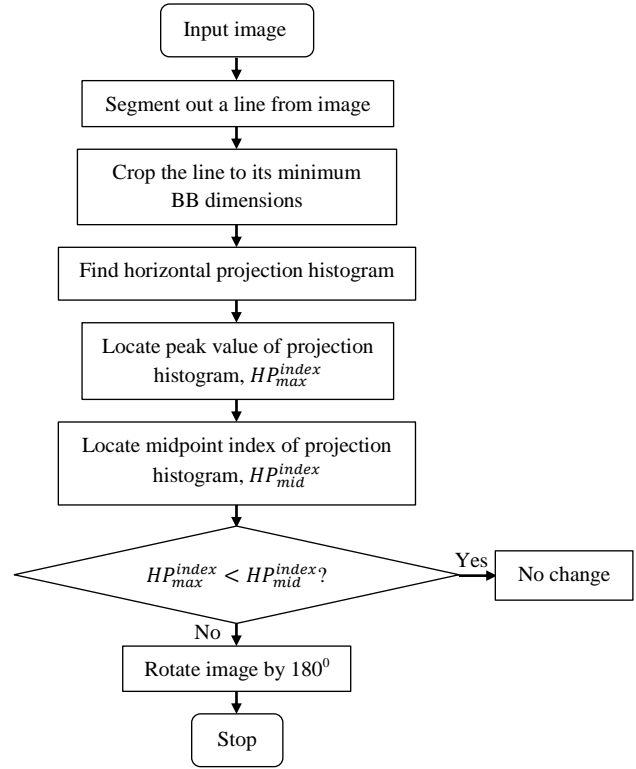

**Fig. 13** Flowchart to correctly align flipped image

3. Find the point $HP_{max}^{index}$ using Eq. (21).

$$HP_{max}^{index} = \max_{m \in [1,H]} HP_L(m) \tag{21}$$

4. If $HP_{max}^{index}$ is less than the midpoint index (say $HP_{mid}^{index}$) of horizontal projection, image is correctly aligned and no change is made. Otherwise the obtained image is flipped upside down and is rotated by $\theta_{skew} = 180$ degrees to obtain final image.

$$HP_{mid}^{index} = \left\lfloor \frac{H}{2} \right\rfloor \tag{22}$$

Where $H$ gives length of horizontal projection and $\lfloor \cdot \rfloor$ denotes greatest integer function.

$$\theta_{skew} = 180° \Leftrightarrow HP_{max}^{index} > HP_{mid}^{index} \tag{23}$$

## 6. Word segmentation

To separate the words, vertical projection of a segmented line say $L_{K \times P}$ is calculated using Eq. (24). The columns for which $I_{VP}(n)$ is zero act as partition between adjacent words as demonstrated in Fig. 14.

$$I_{VP}(n) = \sum_{m=1}^{K} L(m,n) \ \forall \ n \in [1, P] \tag{24}$$

A word subimage $W_{U \times V}$ can be simply represented as combination of different constituent subimages :

<Devanagari word: $W_{U \times V}$> = <Header line: $\overline{H}$> + <Character set with modifiers: $W'$>

<Character set with modifiers: $W'$> = <Ascenders: $W_A$> + < Descenders: $W_D$> + < Individual characters: $\dot{W}$>

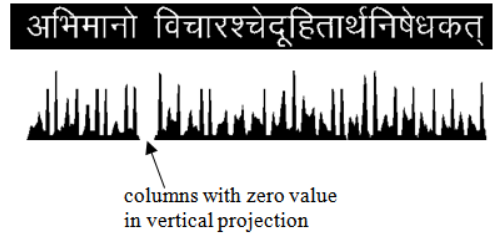

**Fig. 14** Segmented line with its vertical projection showing the columns used for word segmentation

## 7. Character segmentation

Character segmentation process extracts constituent characters from Devanagari word subimage and utilizes following operations for this purpose.

*Vertical Projection*: For a word subimage $W_{U \times V}$, vertical projection $VP_W(n)$ is given by

$$VP_W(n) = \sum_{m=1}^{U} W(m,n) \ \forall \ n \in [1, V] \tag{25}$$

*Horizontal Projection*: Horizontal projection $HP_W(m)$ is found by traversing entire range of rows from $m=1$ to $m=U$ and calculating number of white pixels for each row using Eq. (26).

$$HP_W(n) = \sum_{m=1}^{V} W(m,n) \;\forall\, m \in [1,H] \qquad (26)$$

The above operations are employed few times when performing preliminary character segmentation[4]. Preliminary character segmentation comprises of following steps:

1. Removing the header line: The characters of Devanagari word are connected together via header line. To isolate the characters of a word header line needs to be eliminated. To locate the header line ($\bar{H}$) we compute horizontal projection $HP_W(m)$ of corresponding word and the rows corresponding to highest $HP_W(m)$ value contain the header line as shown in Fig.15(a). Header line can be computed using Eq. (27):

$$\bar{H} = \max_{m \in [1,U]} HP_W(m) \qquad (27)$$

Setting the value of these rows to zero essentially removes the header line. Word subimage without header line, $W'$ can be computed by using Eq. (28):

$$W'_{U \times V} = [W_{U \times V} \mid W_{U \times V}(m,n) = 0 \;\forall m \in \bar{H}, n \in [1,V]] \qquad (28)$$

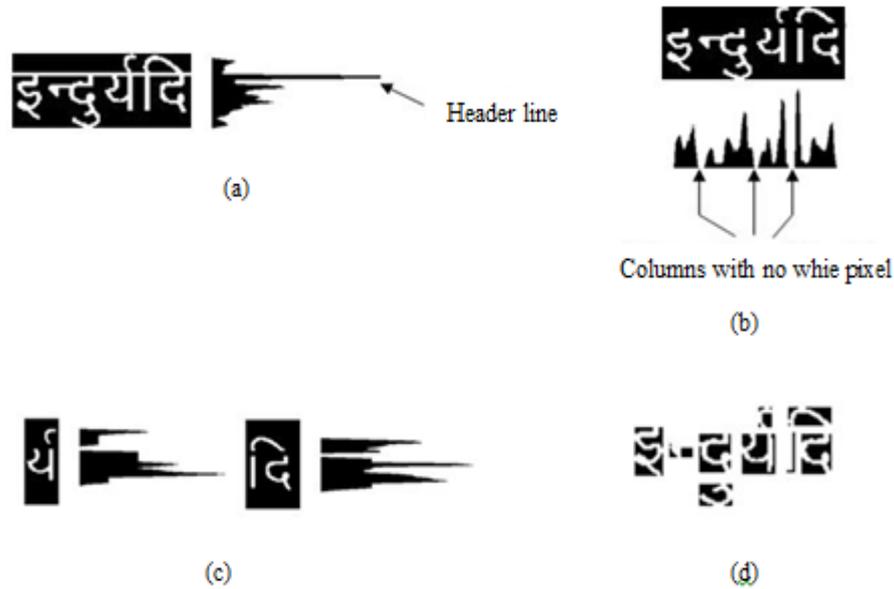

**Fig. 15** Segmentation of a word (a) word image with horizontal projection of word taken to locate header line (b) word image after removing header line with vertical projection to locate segmenting columns (c) segmented core characters with horizontal projection (d) final word after segmentation

2. Separating the characters: After removing header line, residual image $W'_{UxV}$ encompasses core strip fused together with descenders and isolated ascenders. The columns[4] with zero value in vertical projection, given by Eq. (25), are used as delimiters for extracting character subimages from $W'_{UxV}$ as shown in Fig. 15(b).

3. Separating the top modifiers/ascenders: Removing header line delineates upper strip from rest of the image due to the presence of extending black strip as shown in Fig. 15(c), and are easily separated using rows with zero value in horizontal projection using Eq. (26). Result of preliminary segmented word is shown in Fig. 15(d).

*7.1. Segmentation of lower modifiers/ descenders*

Following preliminary character segmentation in previous step, lower modifiers present below the core strip are removed in this section.

Devanagari script has characters of varying height. During preliminary segmentation the end bar from the following characters is removed due to absence of ON (white) pixels in intermediate columns, hence giving us half characters with their height quite less than that of the average height of characters (denoted by *Avg_Ht*) as shown in Fig. 16(a). Then there are characters with their height approximately equal to *Avg _Ht* shown in Fig. 16(b) and characters with lower modifiers with their height greater than *Avg_Ht* as shown in Fig. 16(c). So image boxes with height either greater than *Thresh_Ht* (found using Eq. (29)) or *Avg_Ht* are marked for further segmentation. We use former threshold value because when using *Avg_Ht* as a deciding factor, long character like ह with height slightly greater than *Avg_Ht* is also cropped. This lower cropped portion resembles the lower modifier '*halant*' ( ) very closely. To avoid such undesired segmentation we choose *Thresh_Ht* as threshold value to distinguish the characters for further segmentation.

When a descender is placed below a core character, one of two possible cases occur:

1. Gap between character and modifier: In some cases modifier do not touch the core character as depicted in Fig. 16(c). In such cases segmentation of lower modifier is an easy task and can be accomplished by taking horizontal projection of the character using Eq. (4) and using rows with zero value to separate the modifier.

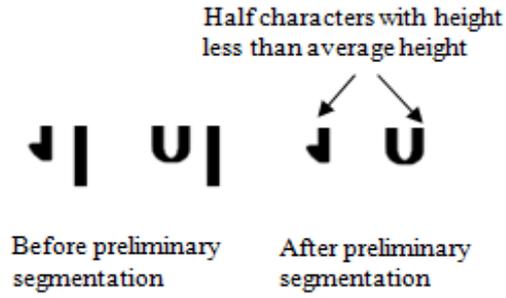

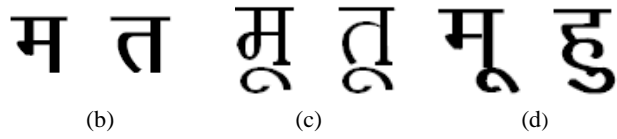

**Fig. 16** Example of devanagari characters with varying height (a) Half characters with height less than $Avg\_Ht$ (b) Character with height equal to $Avg\_Ht$ (c) Height greater than $Avg\_Ht$ with gap between core character and modifier (d) ) Characters with height greater than $Avg\_Ht$ and joined core character and modifier

2. Joined character and modifier: In other cases modifiers are attached to the character below the core strip shown in Fig.16(d). In such cases *Thresh_Ht* is used for segmentation. Detailed discussion is given in following subsection 7.1.1.

*7.1.1. Segmentation of characters with joined lower modifier*

To locate an appropriate row that separates the core character from descender in image $I_D$ (Fig17(b)), we utilise the maximum height of characters. On the basis of threshold found using Eq. (29), segmentation region $R_s$ is evaluated using Eq. (30) and is shown in Fig. 17(b). Segmentation region usually incorporates few rows adjacent to the *Thresh_Ht*. Traversing through all the rows in this region we note the column indices that contain first (leftmost) and last (rightmost) pixel respectively shown in Fig. 18(c). The row say *Seg_row*, with minimum difference between selected *Col_right* and *Col_left* is finalised as segmentation row for separating core character and lower modifier shown in Fig. 18(c).

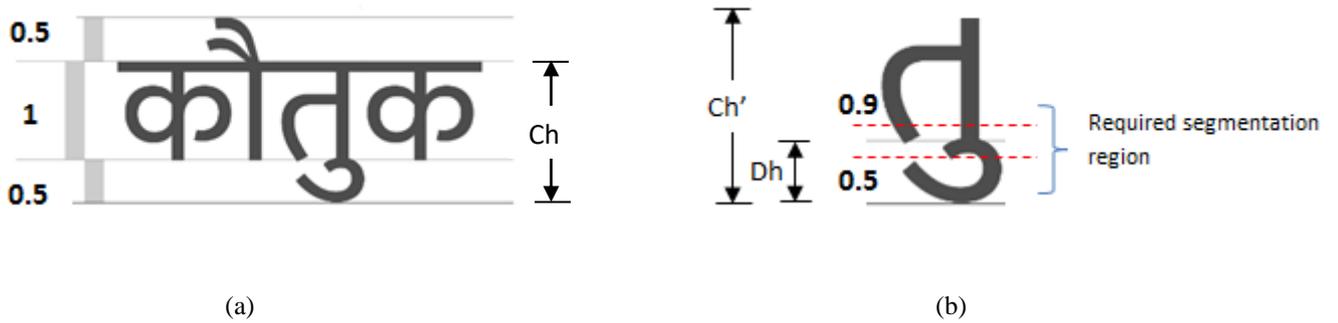

(a)                                                            (b)

**Fig. 17** (a) Standard vertical matra proportion before segmentation[24] (b) Vertical matra proportion before segmentation.

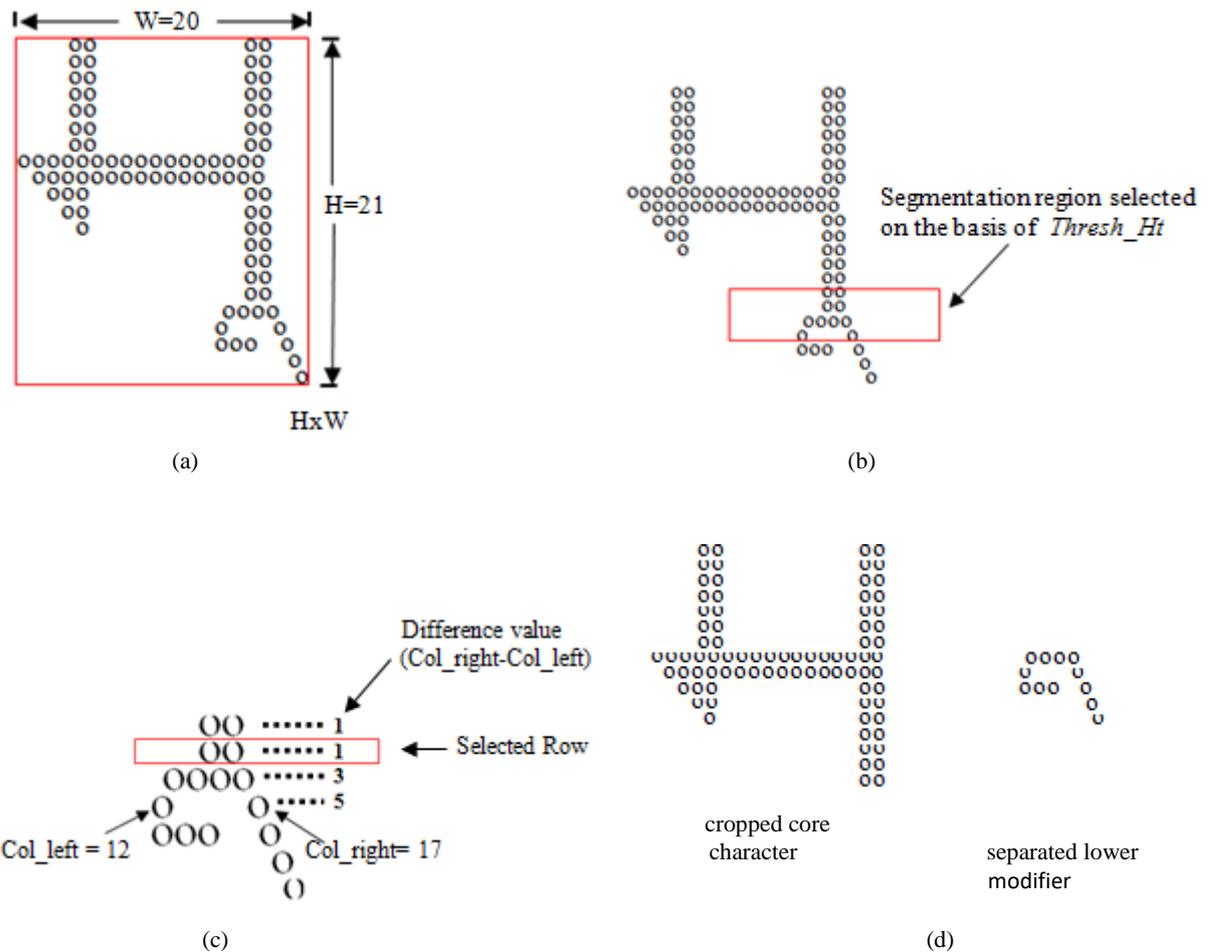

**Fig. 18** Segmentation of characters with joined lower modifier (a) Image $I_D$ of character consisting of lower modifier with dimensions HxW (b) character image with segmentation region within the rectangle (c) Difference value calculated for each row in segmentation region (d) Separated subimages of core character and lower modifier.

$$Thresh\_Ht = \left(1 - {Ch'}/{Ch' + Dh}\right) * (Ch' + Dh) \tag{29}$$

$$R_s = Thresh_{Ht} \pm 10\% \text{ of } Thresh\_Ht \tag{30}$$

$$C^x = \{y | I_D(x,y) = 1\ ;\ \forall y, x \in R_s\}_n \tag{31}$$

$$Col\_left^x = C^x(1)\ \forall\ x \in R_s \tag{32}$$

$$Col\_right^x = C^x(n)\ \forall\ x \in R_s \tag{33}$$

$$Seg\_row = \min_{x \in R_s} Col\_right^x - Col\_left^x \tag{34}$$

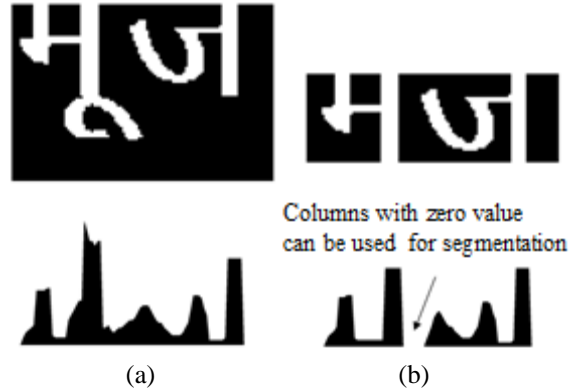

**Fig. 19** (a) Vertical projection of character image before removing lower modifier (b) Vertical projection of character image after removing lower modifier

After removing lower modifiers from core characters, next step involves segmenting those core characters that although are not touching each other but still they cannot be segmented due to absence of valid column with no character pixel. The presence of descender interferes with separation of such non-touching core characters shown in Fig. 19(a). These character subimages can be easily segmented using vertical projection once lower modifier have been removed as shown in Fig. 19(b).

*7.2. Segmentation of conjuncts/touching characters and shadow characters*

Conjuncts in Devanagari script are formed when two or more consonants are joined together usually by removing the right portion of former consonant and affixing it next to an intact consonant shown in Fig.

20(a) and (b). Shadow characters, however, are those characters which do not touch but overlap one another in such a way that they cannot be segmented without clipping off a portion of either character shown in Fig. 20(c) and (d). After preliminary character segmentation and removal of lower modifiers, next step involves separation of touching and shadow characters into their constituent character subimages. The character images marked for further segmentation, due to their width, may either contain conjuncts, shadow characters or both.

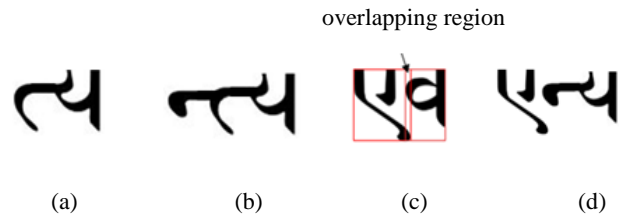

(a)      (b)      (c)      (d)

**Fig. 20** Conjuncts and shadow characters: (a) Conjunct with combination of two characters (b) Conjunct with combination of three characters (c) Example of shadow characters (d) Image with characters in shadow and touching each other.

Primary feature used to separate conjuncts or shadow character is width of the character. Decisive threshold to select characters for further segmentation is based on average width of characters and is chosen based on following observations regarding conjuncts and shadow characters:

1. The width of composite characters is comparable to or greater than twice the average width ($Avg\_Wd$) of characters.

2. The width of a conjunct composed of two constituent Devanagari characters is within close range to twice the average width. Fig. 21(b) shows example of such conjuncts. Threshold range for such conjuncts is chosen using Eq. (35):

$$Thresh\_Wd_2 = 2 * Avg\_Wd : 3 * Avg\_Wd \qquad (35)$$

3. The width of a conjunct comprising three constituent Devanagari characters is usually three times the average width. Threshold width chosen for such conjuncts is given by Eq. (36). Figure 21(c) shows example of conjuncts consisting 3 characters joined together.

$$Thresh\_Wd_3 > 3 * Avg\_Wd \qquad (36)$$

4. For composite character image consisting of two characters in shadow or two characters touching each other, the segmentation region is usually the mid region with approximate range equivalent to $R_s^1$ given by Eq. (37).

$$R_s^1 = Avg\_Wd \pm 20\% \text{ of } Avg\_Wd \tag{37}$$

5. For composite character image consisting of three character subimages, the first segmenting column lies in region $R_s^1$ and second segmenting column lies in region $R_s^2$ approximately near the twice of $Avg\_Wd$, Eq. (38).

$$R_s^2 = 2 * Avg\_Wd \pm 20\% \text{ of } Avg\_Wd \tag{38}$$

Using above observations, the resultant segmenting column, required to separate the constituent characters, is calculated.

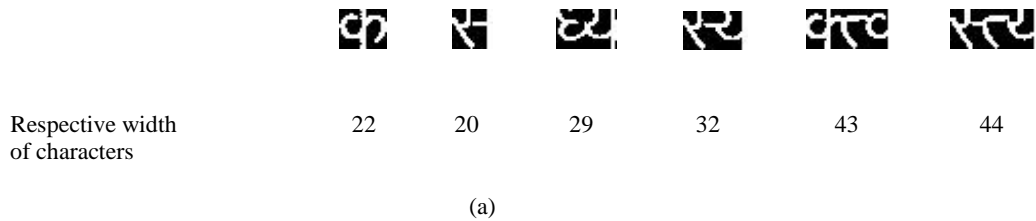

| Respective width of characters | 22 | 20 | 29 | 32 | 43 | 44 |

(a)

Mean width (as calculated while processing whole document) = 12
Width range for conjuncts with 2 characters = 24: 36
Width for conjuncts with 3 characters > 36

Characters selected for further segmentation based on above data:

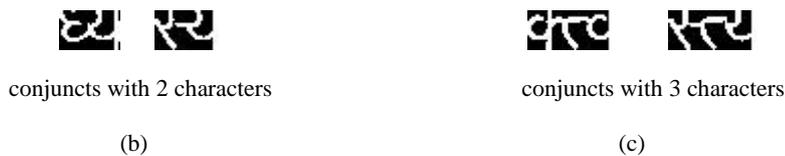

conjuncts with 2 characters           conjuncts with 3 characters

(b)                                   (c)

**Fig. 21**   Analysing conjuncts on the basis of their width (a) characters with their respective width (b) conjuncts with 2 characters   (c) conjuncts with 3 characters

*7.2.1. Segmentation of conjuncts*

Devanagari characters with their width satisfying Eq. (35) and (36) are marked for further segmentation. The appropriate column for cropping out the constituent character subimages is found by applying the following algorithm, within the segmentation region of a composite character. The corresponding segmentation region for composite characters is mentioned in points 4 and 5 of preceding section.

1. Scan each of the character image to mark the rows that contain first and last white pixel using Eq. (39)-(41).

$$Row^y = \{x | I_D(x,y) = 1 \,;\, \forall x, y \in R_s^i\}_n \tag{39}$$

$$Row\_Top^y = Row^y(1) \,\forall\, y \in R_s^i \tag{40}$$

$$Row\_Bottom^y = Row^y(n) \,\forall\, y \in R_s^i \tag{41}$$

2. Next for each column, difference between lowermost and uppermost row is calculated.

3. The column with minimum difference value is chosen for segmentation as shown in Fig. 22(b).

$$Seg\_col = \min_{y \in R_s^i} (Row\_Bottom^x - Row\_Top^x) \tag{42}$$

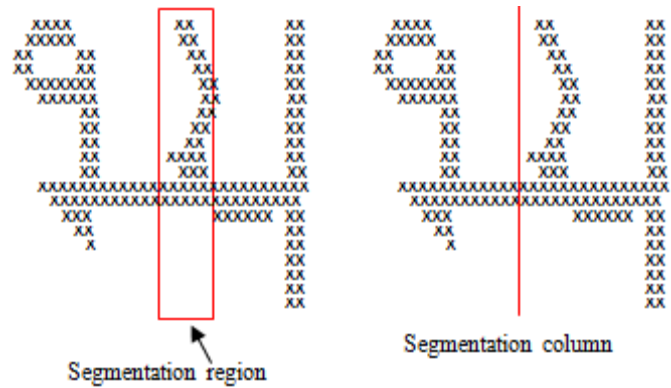

Fig. 22 The image of conjunct: (a) Segmentation region in conjunct character (b) Image with resulting segmentation column

For conjuncts made up of two characters, a single segmenting column is required.

To determine segmenting column for such conjuncts Eq. (39)-(42) are used with $R_s^i$ replaced with $R_s^1$.

For conjuncts comprising three core characters first segmenting column is found using Eq. (39)-(42) with $R_s^i$ replaced with $R_s^1$ whereas for second segmenting column $R_s^i$ replaced with $R_s^2$ in Eq. (39)-(42).

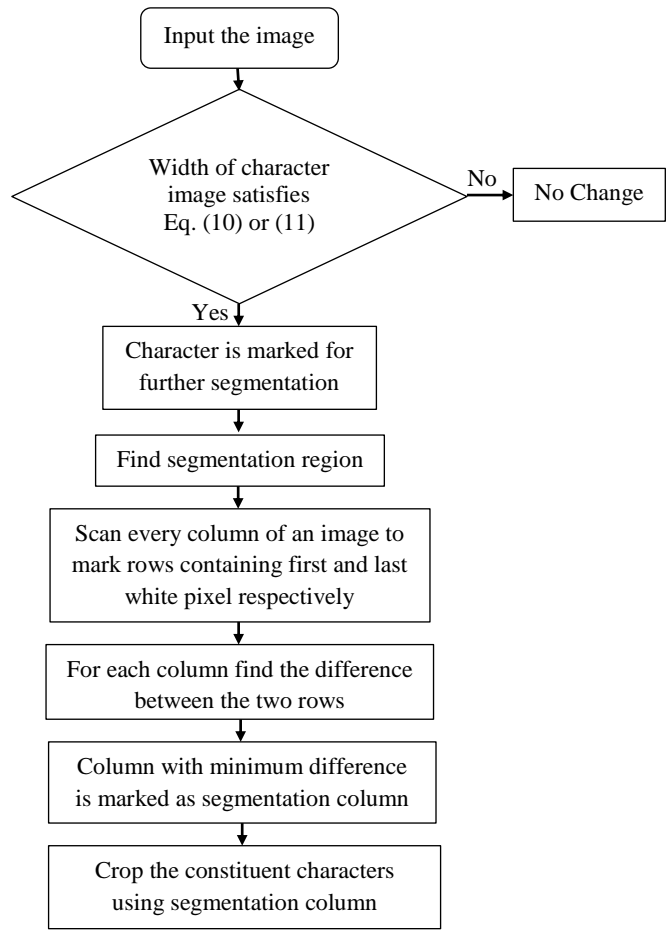

**Fig. 23** Flowchart for segmenting conjuncts.

*7.3. Segmentation of shadow characters*

Previous works of segmenting shadow characters[4] find an appropriate segmenting column to separate the constituent characters. When using segmentation column to separate shadow characters, a little portion of either character is clipped off. Segmentation of shadow characters, using algorithm given in Sec. 7.2.1., is illustrated in Fig. 24. As seen in Fig. 24(d) segmentation crops a small portion from first character. To avoid such errors we have used an algorithm based on CC, as depicted in flowchart of Fig. 26, to isolate shadow characters.

Before applying CC algorithm to separate shadow characters, image is closed using a square SE. This action fills any small gap present in constituent character subimages as demonstrated in Fig. 25.

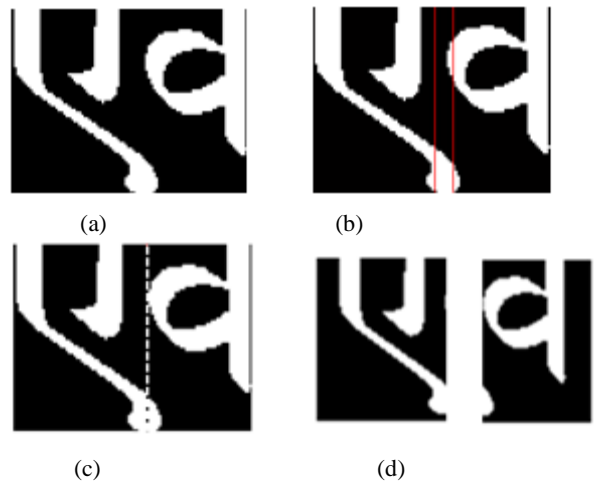

**Fig. 24** Characters in shadow and their segmentation: (a) Characters in shadow (b) Image showing segmentation region within the rectangle (c) Segmentation column represented with dotted line (d) Individual characters after segmentation

Resulting dilated image has number of 8-connected components equivalent to total number of core characters. Image 25(a) shows two characters in shadow with each other with three CCs. Dilated image, in Fig. 25(b), has two core characters corresponding to two CCs.

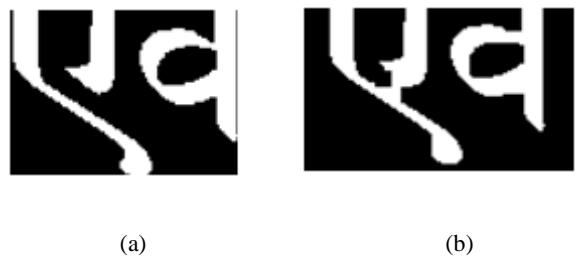

**Fig. 25** (a) Example of shadow character with gap in constituent character shown with dotted circle (b) dilated image with number of core characters corresponding to number of CCs

Resulting dilated image number of CCs are calculated and corresponding pixels of one component are set to zero as shown in Fig. 27(a). Subtracting this image from original gives an isolated character that was removed previously and further subtracting this isolated character from original image results in subimage of another constituent character.

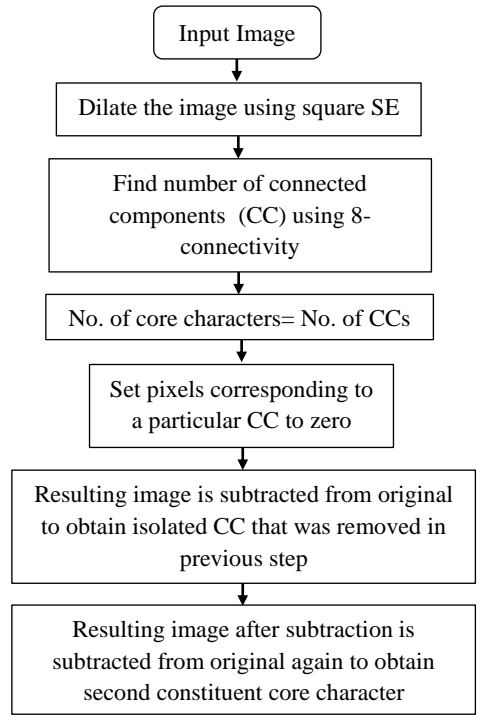

**Fig. 26** Flowchart representing method to isolate shadow characters

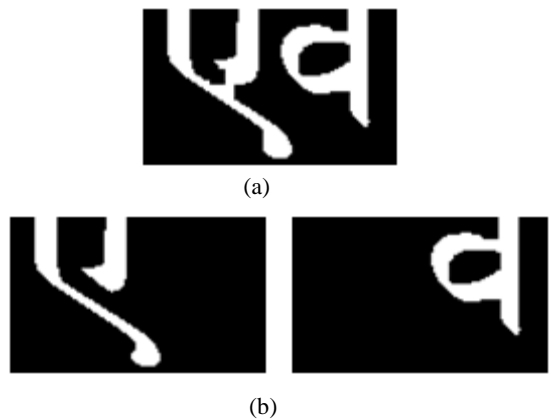

**Fig. 27** (a) Dilated image with shadow character (b) constituents characters obtained after subtraction

## 8. Feature extraction and classification of modifiers

To distinguish different modifiers from one another, the modifiers are skeletonized down to unitary thickness before extracting features using 8-pixel adjacency. For each non-zero pixel $M_{IxQ}(r,c)$ (Fig. 28) in a skeletonized modifier subimage, number of foreground neighbouring pixels (*NP*) can be found using Eq. (43).

$$NP = \sum_{c=c-1}^{c+1} \sum_{r=r-1}^{r+1} M(r,c) - 1 | M(r,c) = 1 \tag{43}$$

| M(r-1,c-1) | M(r-1,c) | M(r-1,c+1) |
| --- | --- | --- |
| M(r,c-1) | M(r,c) | M(r,c+1) |
| M(r+1,c-1) | M(r+1,c) | M(r+1,c+1) |

**Fig. 28** Example of 8-connected neighbours with origin at *M(r,c)*

Extreme ends of skeletonized modifiers are the pixels with just one 8-adjacent element and are calculated using following Eq. (44):

$$M_{end}(r,c) = \{M(r,c) | NP = 1\} \tag{44}$$

For such pixels following features are calculated using Eq. (45)-(48):

$$\text{Left\_neighbours} = \sum_{r=r-1}^{r+1} M(r, c-1) \tag{45}$$

$$\text{Right\_neighbours} = \sum_{r=r-1}^{r+1} M(r, c+1) \tag{46}$$

$$\text{Top\_neighbours} = \sum_{c=c-1}^{c+1} M(r-1, c) \tag{47}$$

$$\text{Bottom\_neighbours} = \sum_{c=c-1}^{c+1} M(r+1, c) \tag{48}$$

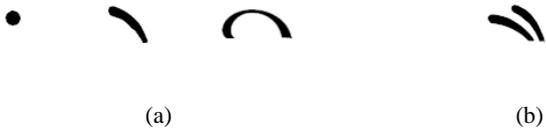
          (a)                            (b)

**Fig. 29** Top strip components (a) with one CC (b) with two CCs

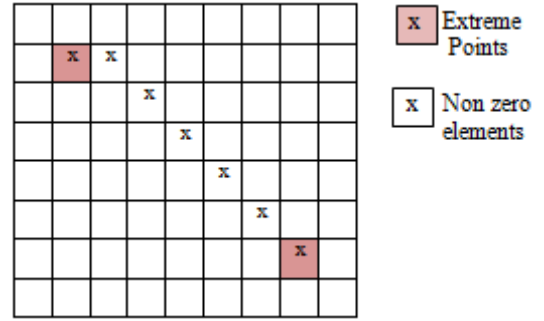

**Fig. 30** Modifier image showing extreme points

All of the upper-strip elements have one CC, except for ( ` ) as shown in Fig. 29, and hence it can easily be recognised from all other modifiers with single CC. The modifiers with a single CC have two extreme points as shown in Fig. 30. For both of these pixels respective features given by Eq. (45)-(48) are used to differentiate between modifiers. An example illustrating how these features are used to differentiate between different modifiers is shown in Fig. 31.

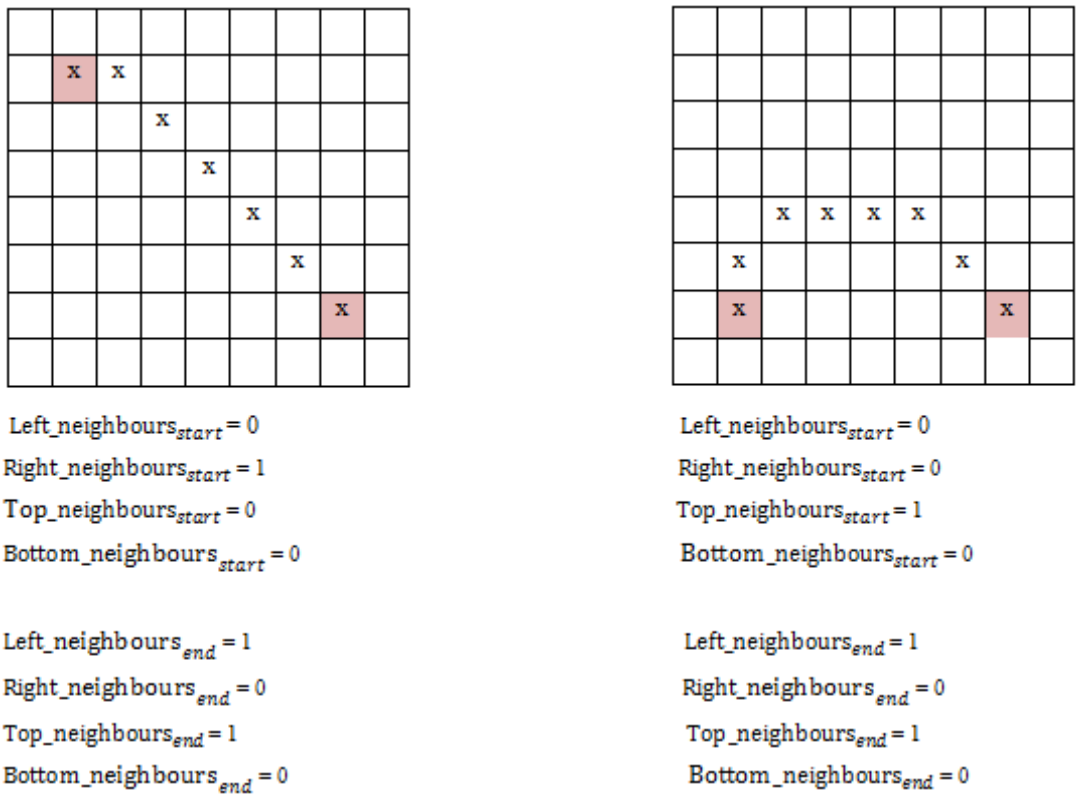

**Fig. 31** Example illustrating neighbouring features of two different modifiers.

## 9. Feature extraction for character recognition

In the present section we discuss features which will be utilized to recognize Devanagari characters.

**Zoning:** To increase the character recognition accuracy, the image is partitioned into non-overlapping non-uniform regions zones as shown in Fig. 32 and number of character pixels are calculated for each of these zones.

To find number of pixels in a particular zone, a mask is formed such that all the coefficients outside respective zone are zero as shown in Fig 33. For image $I_{MxN}$ with corresponding mask $w$ of same size, number of foreground pixels ($NOP_i$) in $i^{th}$ region is given by Eq. (49).

$$NOP_i = \sum_{p=1}^{M} \sum_{q=1}^{N} I(p,q) * w_i(p,q) \qquad (49)$$

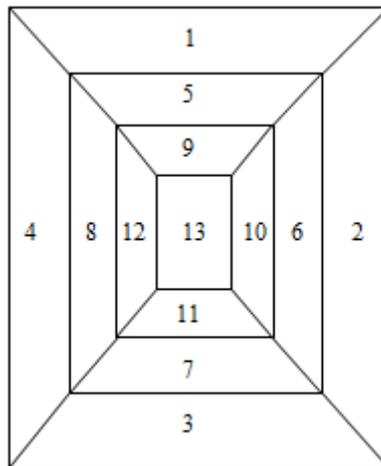

**Fig. 32** Grid showing 13 different non-uniform zones

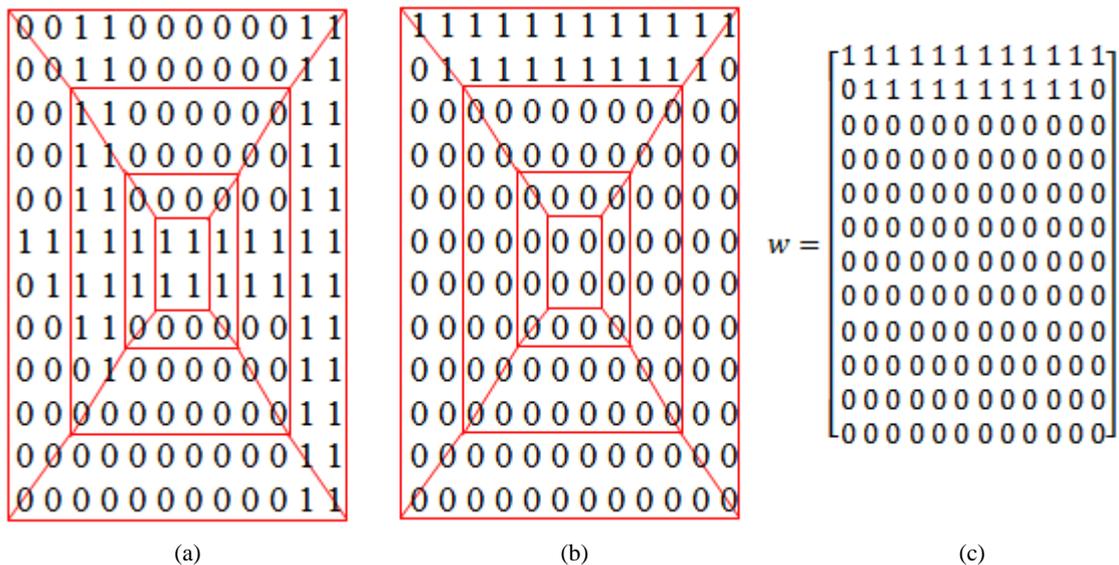

(a)           (b)           (c)

**Fig. 33** (a) Character image partitioned in different zones (b),(c) mask corresponding to region

**Crossings** : Crossings depicts number of transitions from foreground to background pixels along columns and rows in a character image $I_{MxN}$

Horizontal transitions:

$$HT(m) = card(\,I(m,n)|I(m,n) = 1 \land I(m,n+1) = 0\,; \quad \forall\, m \in M, n \in N-1 \tag{50}$$

Vertical transitions:

$$VT(n) = card(\,I(m,n)|I(m,n) = 1 \land I(m+1,n) = 0\,; \quad \forall\, m \in M-1, n \in \tag{51}$$

**Projection histograms**: Projection histograms compute number of foreground pixels in each row and column of character image $I_{MxN}$.

Horizontal projection histogram

$$HP(m) = \sum_{n=1}^{N} I(m,n) \;\forall\, m \in [1, M] \tag{52}$$

Vertical projection histogram

$$VP(n) = \sum_{m=1}^{M} I(m,n) \;\forall\, n \in [1, N] \tag{53}$$

**Moments**: For projection and transition features $NOP_i$, $HT(m), VT(n), HP(m), VP(n)$, various geometrical moments are found using formulas given in the table 1.

Table 1 Formulas for calculating different transition and projection moments

|  | Transition moments | Projection moments |
|---|---|---|
| Mean | Horizontal transition mean $$\overline{HT} = \frac{1}{M} \sum_{j=1}^{M} HT_j$$ Vertical transition mean $$\overline{VT} = \frac{1}{N} \sum_{j=1}^{N} VT_j$$ | Horizontal projection mean $$\overline{HP} = \frac{1}{M} \sum_{j=1}^{M} HP_j$$ Vertical projection mean $$\overline{VP} = \frac{1}{N} \sum_{j=1}^{N} VP_j$$ |
| Variance | Horizontal transition variance $$Var_{HT} = \frac{1}{M}\sum_{j=1}^{M}(HT_j - \overline{HT})^2$$ Vertical transition variance $$Var_{VT} = \frac{1}{N}\sum_{j=1}^{N}(VT_j - \overline{VT})^2$$ | Horizontal projection variance $$Var_{HP} = \frac{1}{M}\sum_{j=1}^{M}(HP_j - \overline{HP})^2$$ Vertical projection variance $$Var_{VP} = \frac{1}{N}\sum_{j=1}^{N}(VP_j - \overline{VP})^2$$ |
| Skew | Horizontal transition skew $$Skew_{HT} = \frac{1}{M}\sum_{j=1}^{M}\left(\frac{HT_j - \overline{HT}}{\sigma_{HT}}\right)^3$$ Vertical transition skew $$Skew_{VT} = \frac{1}{N}\sum_{j=1}^{N}\left(\frac{VT_j - \overline{VT}}{\sigma_{VT}}\right)^3$$ | Horizontal projection skew $$Skew_{HP} = \frac{1}{M}\sum_{j=1}^{M}\left(\frac{HP_j - \overline{HP}}{\sigma_{HP}}\right)^3$$ Vertical projection skew $$Skew_{VP} = \frac{1}{N}\sum_{j=1}^{N}\left(\frac{VP_j - \overline{VP}}{\sigma_{VP}}\right)^3$$ |
| Kurtosis | Horizontal transition kurtosis $$Kurt_{HT} = \left\{\frac{1}{M}\sum_{j=1}^{M}\left(\frac{HT_j - \overline{HT}}{\sigma_{HT}}\right)^4\right\} - 3$$ Vertical transition kurtosis $$Kurt_{VT} = \left\{\frac{1}{N}\sum_{j=1}^{N}\left(\frac{VT_j - \overline{VT}}{\sigma_{VT}}\right)^4\right\} - 3$$ | Horizontal projection kurtosis $$Kurt_{HP} = \left\{\frac{1}{M}\sum_{j=1}^{M}\left(\frac{HP_j - \overline{HP}}{\sigma_{HP}}\right)^4\right\} - 3$$ Vertical projection kurtosis $$Kurt_{VP} = \left\{\frac{1}{N}\sum_{j=1}^{N}\left(\frac{VP_j - \overline{VP}}{\sigma_{VP}}\right)^4\right\} - 3$$ |

## 10. Recognition

Recognition stage comprises of two phases. In the first phase of recognition, library character samples are sorted according to minimum Euclidean distance between library feature vector and input feature vector. For the first phase, feature vector $S_{reg}$ comprises of 13 features each corresponding to number of foreground pixels in $i^{th}$ region (Eq. (49)) and $S_{reg}^i$ represents $i^{th}$ feature in the corresponding feature vector $S_{reg}$.

$$S_{reg} = [NOP_1\ NOP_2\ NOP_3\ NOP_4\ NOP_5\ NOP_6\ NOP_7\ NOP_8\ NOP_9\ NOP_{10}\ NOP_{11}\ NOP_{12}\ NOP_{13}] \quad (54)$$

Euclidean distance between library feature vector $S_{reg}^i$ and input feature vector $X_{reg}^i$ can be calculated using Eq. (55).

$$D_{reg} = \sqrt{\sum_{i=1}^{13}(S_{reg}^i - X_{reg}^i)^2} \quad (55)$$

Where $S_{reg}^i$ is $i^{th}$ region feature for database character sample and $X_{reg}^i$ is ith input feature.

For the second phase of recognition sorted database character samples are selected with minimum Euclidean distance and from these selected samples nearest neighbour to input character is found. To find nearest neighbour Euclidean distance between input feature vector and library feature vector Eq. (56) is used.

$$D_{mom} = \sqrt{\sum_{i=1}^{16}(S_{mom}^i - X_{mom}^i)^2} \quad (56)$$

Where $S_{mom}$ is feature vector containing all the moments given in table 1 and $S_{mom}^i$ represents $i^{th}$ feature in corresponding vector. $X_{mom}^i$ is $i^{th}$ moment feature in input feature vector $X_{mom}$.

Input character is matched with the database that has minimum distance, $D_{mom}$.

*10.1. Inadequate classification due to similarity in character pairs after removing header line*

In some cases removing complete header line clips off significant part of a character and hence making it appear similar to another character. Some character pairs, shown in Fig. 34(a), differ only in header line region. After removing header line, it becomes difficult to distinguish these characters, causing classification errors as seen in Fig. 34(b).

To distinguish these characters after removing header line, we compute number of CCs and on basis of number of CCs we classify them in the correct group.

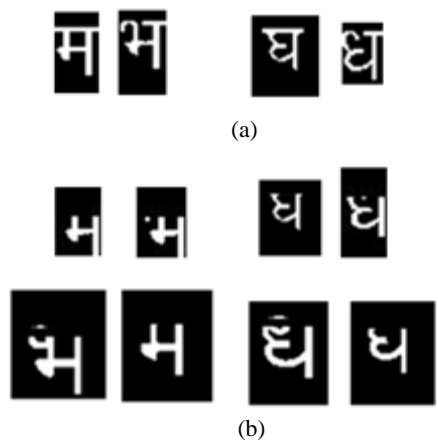

**Fig. 34** (a) Character pairs with similar structure except header line portion (b) indistinguishable character pairs after removing header line

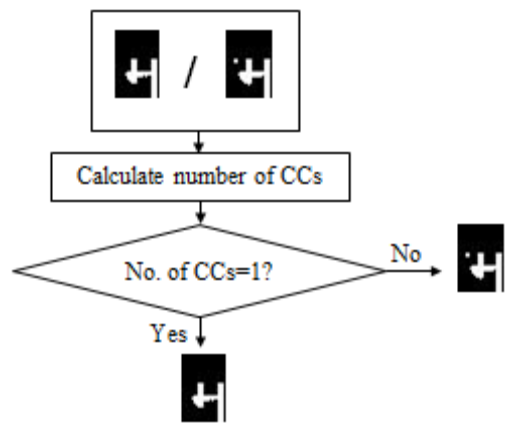

**Fig. 35** Flowchart distinguishing between similar looking characters after header line removal.

*10.2. Inadequate character classification due to rakar modifier*

Combination of a consonant and rakar modifier goes unnoticed by segmentation algorithm because the resulting composite character has similar dimensions as any other basic core character. Few examples of consonant-rakar combination are shown in Fig. 36. Probability of occurrence of such characters is far too minimal to have any considerable effect on the accuracy of proposed algorithm except for प्र. In transliteration algorithm, the character प्र is mapped to प in most of the cases. We use number of end points as a decisive feature to accurately classify these characters as shown in Fig. 37. For skeletonized character image, number of end points is computed using Eq. (44).

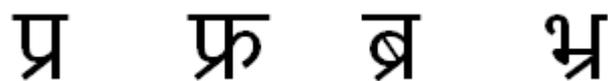

**Fig. 36** Few examples of consonant-rakar combination

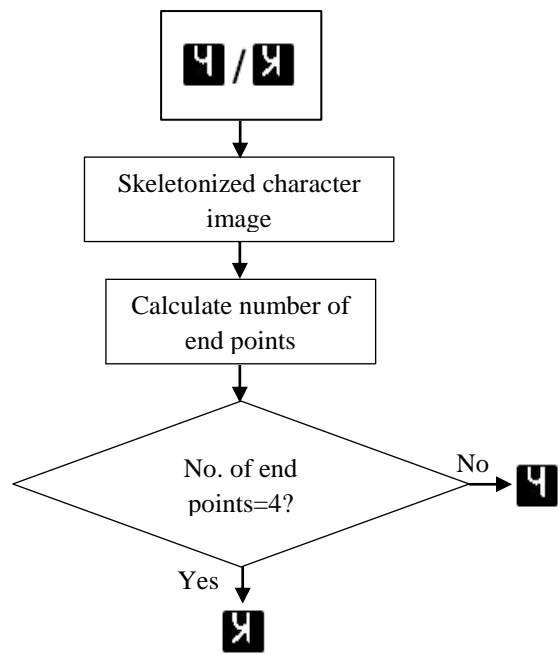

**Fig. 37** Flowchart for accurate classification of a consonant–rakar combination

## 11. Transliteration

The mapping between Devanagari and roman script character in a transliteration scheme should be as close as possible to the pronunciation of the source character in the target script. We have employed

International Alphabet of Sanskrit Transliteration (IAST) scheme for romanization of recognized Devanagari characters. Figure 38 shows phonetically similar roman alphabets for each Devanagari character.

**12. Results**

We have tested the proposed method for several Devanagari documents. Some of these documents are available at http://sanskrit.safire.com/Sanskrit.html#Saptashati. The proposed algorithm recognizes the characters extracted from corresponding Devanagari document image and results in a text file containing Roman character corresponding to each recognized Devanagari character. The proposed algorithm has been implemented using Matlab R2014a programming. Results of character segmentation of sample text and roman transliteration after recognition are shown in following Fig. 39. and Fig. 40

Following two points aid in interpreting the given transliteration results:

i) The characters underlined in red represent true characters that are not first choice.
ii) Underscore in transliteration result represents a modifier that has not been recognized.

| | |
|---|---|
| अ आ इ ई उ ऊ | |
| a ā i ī u ū | |

ए ऐ ओ औ अं अः
e ai o au aṁ aḥ

ऋ ॠ ऌ ॡ
ṛ ṝ ḷ ḹ

ं ः ँ
ṁ ḥ ☐

क ख ग घ ङ
ka kha ga gha ṅa

च छ ज झ ञ
ca cha ja jha ña

ट ठ ड ढ ण
ṭa ṭha ḍa ḍha ṇa

त थ द ध न
ta tha da dha na

प फ ब भ म
pa pha ba bha ma

य र ल व
ya ra la va

श ष स ह
śa ṣa sa ha

**Fig. 38** Phonetically similar Roman alphabet for each Devanagari character

(a)

(b)

(c)

**Fig. 39** (a) Input binary Devanagari image (b) Character subimages after preliminary segmentation (c) Segmentation result after conjunct and shadow character separation

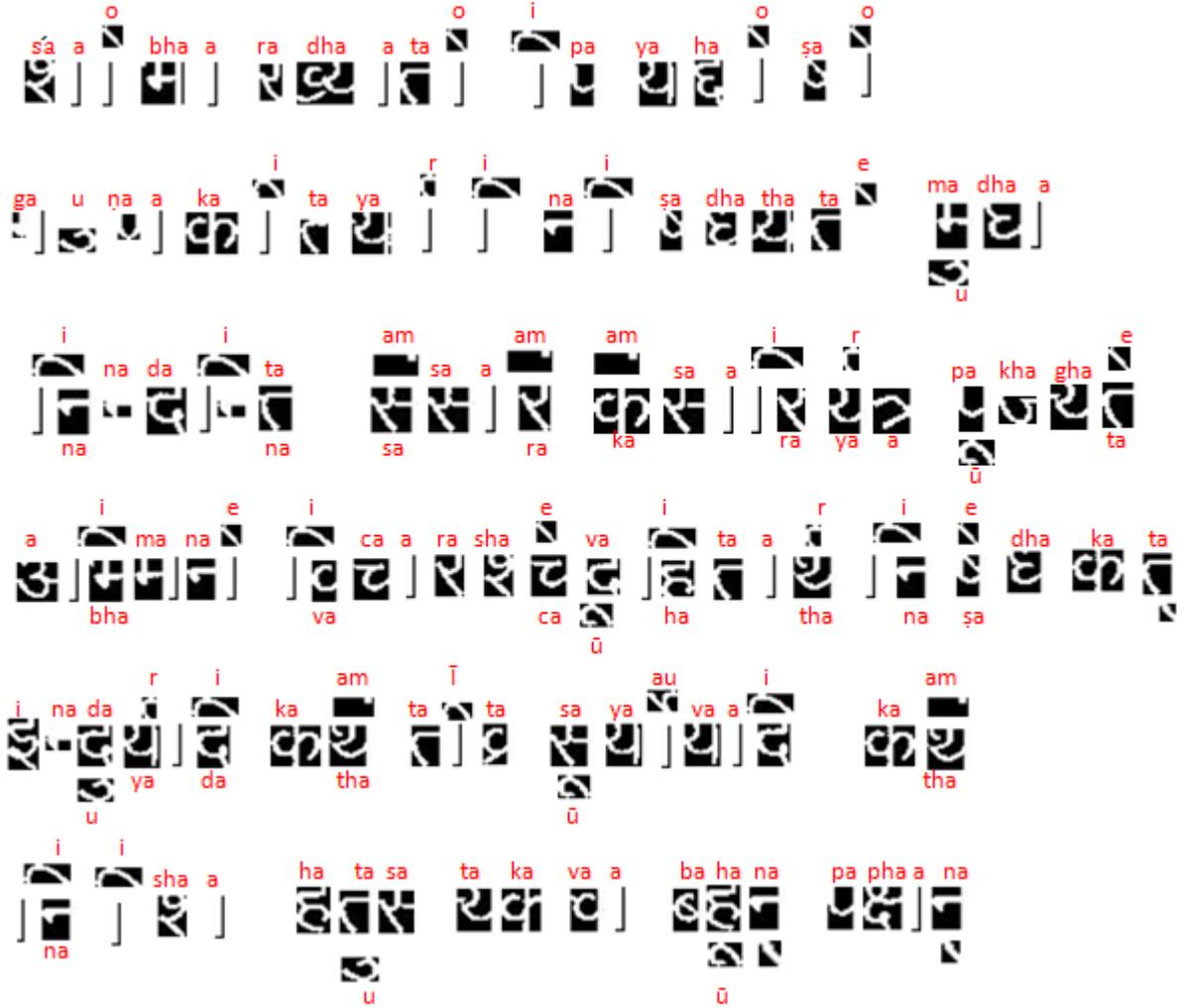

śaaobhaa radhaatao pai yahaoṣa gauṇaakaitayar naiṣaidhathatae

maudhaa nainadanaita saamsaaraam kaamsarairyaa paūkhaghatae

abhaimanae vaicaarashacaevauhaitaarthanaiṣadhakata

inadauryaida kathaam taīta saūyaau vaadai katham

naishaia hatausatakavao bahaūna pakṣaana

**Fig. 40** Result of Romanization of Devanagari characters.

*12.1 Performance Analysis*

To determine effectiveness of our algorithm, we have chosen to analyse results of character segmentation, recognition and transliteration process as the presented algorithm segments the lines and words in the document with 100% accuracy.

The proposed algorithm gives promising results for character segmentation of Devanagari document images and is able to perform preliminary segmentation with 100% accuracy. To evaluate efficiency of character segmentation process in the algorithm, we determine how well the presented algorithm can detect composite characters in an image and how precise are the segmentation cuts for such characters. In order to evaluate the performance of composite character segmentation we use following two parameters:

i) The **precision rate** is defined as ratio of correctly segmented composite characters to the sum of correctly segmented composite characters and false positives. **False positives** (FP) are those characters in the image that although are not composite characters but have been marked for further segmentation.

$$Precison\ rate = \frac{correctly\ segmented\ composite\ characters}{correctly\ segmented\ composite\ characters + FP} \times 100 \quad (57)$$

ii) The **recall rate** is defined as ratio of correctly segmented composite characters to the sum of correctly segmented composite characters and false negatives. **False negatives** (FN) are those characters in the image that although are composite characters but have not been marked for further segmentation.

$$Recall\ rate = \frac{correctly\ segmented\ composite\ characters}{correctly\ segmented\ composite\ characters + FN} \times 100 \quad (58)$$

Table 2 discusses the results of composite character segmentation. The data incorporated in the table has been obtained by executing character segmentation algorithm on several Devanagari documents. Out of total composite characters, proposed algorithm correctly detects about 85% of the characters and correctly segments 95% of detected composite characters. 99% of the composite characters that go undedicated are special cases of consonant combinations as shown in Fig. 41. These special case conjuncts have same height and width as that of a regular core character.

क्+ ष　　ज्+ ञ　　त्+ त　　त्+ र　　द्+ म
　क्ष　　　ज्ञ　　　त्त　　　त्र　　　द्म

**Fig. 41** Special cases of conjunct combinations

Table 2 Performance of composite character segmentation

| Image | Total composite char. | Accurately detected char. | Accurately segmented char. | FP | FN | P % | R % |
|---|---|---|---|---|---|---|---|
| 1 | 10 | 9 | 8 | 1 | 1 | 88.89 | 88.89 |
| 2 | 9 | 7 | 7 | 0 | 2 | 100 | 77.78 |
| 3 | 7 | 5 | 5 | 1 | 2 | 83.33 | 71.42 |
| 4 | 18 | 17 | 17 | 1 | 1 | 94.44 | 94.44 |
| 5 | 10 | 10 | 10 | 0 | 0 | 100 | 100 |
| 6 | 14 | 11 | 11 | 0 | 3 | 100 | 78.57 |
| 7 | 6 | 3 | 3 | 0 | 3 | 100 | 50 |
| 8 | 5 | 3 | 3 | 1 | 2 | 75 | 60 |
| 9 | 5 | 4 | 2 | 4 | 0 | 33.33 | 100 |
| Total | 84 | 69 82.14% | 66 95.56% | 8 | 14 | 89.18 | 82.50 |

Table 3 evaluates the performance of character recognition and transliteration stage. The proposed method gives an overall accuracy of about 90% when some of the true characters are not first choice whereas this accuracy reduces to about 80% when all the true characters in transliterated text are first choice.

Table 4 compares the proposed skew correction methodology with existing algorithms. Performance of proposed method has been tested using Matlab R2014a and timing value corresponds to the time taken by the given algorithm to give final unskewed image. The results of these methods are inferior for any angle that lies outside the corresponding mentioned range. Compared to all these methods, the proposed method works for the full range of angles.

Table 3 Performance of Devanagari transliteration process

| Image | Total char. | Total transliterated char. | Overall accurate transliteration | No. of Romanized char. that are not first choice | Error % |
|---|---|---|---|---|---|
| 1 | 134 | 130 | 122 | 0 | 6.15 |
| 2 | 138 | 136 | 130 | 9 | 4.41 |
| 3 | 129 | 127 | 117 | 7 | 7.87 |
| 4 | 201 | 197 | 181 | 10 | 8.12 |
| 5 | 110 | 108 | 101 | 9 | 6.48 |
| 6 | 147 | 145 | 126 | 6 | 13.10 |
| 7 | 125 | 122 | 110 | 7 | 5.45 |
| 8 | 112 | 105 | 86 | 6 | 18.09 |
| 9 | 147 | 147 | 130 | 7 | 11.56 |
| Total | 1243 | 1217 97.90% | 1103 90.63% | 61 5.53% | 9.36 |

Table 4 Comparison of proposed skew correction methodology

| Methods | Range | Average time (secs) |
|---|---|---|
| Yang Cao[26] | -10°-10° | < 2 |
| A.K. Das [29] | -10°-10° | < 1 |
| S. Li [20] | -15°-15° | < 55 |
| E. Kavallieratou [27] | -89°-89° | < 8 |
| A. Papandreou [19] | -5°-5° | -- |
| M. Shafii [28] | -90°-90° | < 2 |
| Proposed Method | -180°-180° | < 3 |

## 13. Conclusion

In this paper, we presented a complete character recognition system for Devanagari documents followed by transliteration of recognized Devanagari characters to roman script. Proposed method efficiently segments skewed and overlapped lines using CC analysis. Using CCs to separate overlapping lines, keeps the individual lines intact and quickens the process as compared to contouring the individual lines. Proposed skew correction methodology works for full range of angles and is lot more faster than traditional projection profile method cause of reduced skew angle variation range . We have profusely used structural properties for character segmentation process, discussing all the possible scenarios like conjunct and shadow character segmentation. Feature extraction methodology described here uses moment based features calculated for projection histogram and crossings together with zone based features. We have used two phase recognition scheme for classification procedure and discussed header line based errors encountered while recognition process. Recognition process is followed by transliteration of Devanagari characters into corresponding roman alphabets based on phonetic similarities. Efficient automatic transliteration will help in conversion of Devanagari documents making it easier for a person, unfamiliar with Devanagari script, to understand them.

**Jasmine Kaur** received her B. Tech. degree in Electronics and Communication (2014) and M. E. in Wireless Communication from Thapar University in 2016. Her current research interests include image processing.

**Vinay Kumar** is working with Thapar University in Electronics and Communication department as an assistant professor. He had completed his PhD in the field of image and signal processing. His research interests are image and video processing

**Caption list**

**Fig. 1** Character zones in Devanagari script

**Fig. 2** Stages of Devanagari recognition process and errors in different stages.

**Fig. 3** (a) Histogram of input image (b) output binary image with threshold $T$=170 (c) output binary image with threshold $T$=30 (d) output binary image with threshold $T$=220.

**Fig. 4** Line segmentation (a) Input image $I'_{HxW}$ with its horizontal projection (b) Constituent segmented lines.

**Fig. 5** (a) Overlapping lines with horizontal projection (b) segmented lines by drawing a horizontal parallel cut

**Fig. 6** (a) Example of 4-pixel connectivity where pixels P0, P1, P2, P3 and P4 form a 4-pixel connected component (b) One connected component based on 8-pixel connectivity

**Fig. 7** Flowchart representing method to isolate overlapping line

**Fig. 8** (a) Overlapping lines to be segmented (b) isolated modifiers removed (c) corresponding words of a line grouped into one connected component (d) image after removing one connected component (e) segmented line

**Fig. 9** Example of dilation with line SE of varying angles (a) original image (b) dilation with SE at angle of $0^0$ (c) dilation with SE at angle of $5^0$ (d) dilation with SE at angle of $10^0$

**Fig. 10**   Example of closing with SEs of different size (a) original image (b) closing with SE of size 5x5 (c) closing with SE of size 10x10 (d) closing with SE of size 15x15

**Fig. 11** Flowchart depicting skew correction methodology

**Fig. 12**   Skew correction of anticlockwise aligned image (a) original image (b) horizontal projection of original image (c) closed image using square SE of size 15x15 (d) horizontal projection of skewed image (e) image rotated clockwise by $2^0$ (f) horizontal projection of clockwise rotated image with peak value 128 (g) image rotated anticlockwise by $2^0$ (h) horizontal projection of anticlockwise rotated image with peak value 111 (i) image dilated using line SE of size 100 at angle $10^0$ (j) image dilated using line SE of size 100 at angle $20^0$ (k) image dilated using line SE of size 100 at angle $40^0$ (l) image dilated using line SE of size 100 at angle $55^0$ (m)  image corrected by rotating original image by $-46^0$.

**Fig. 13** Flowchart to correctly align flipped image

**Fig. 14** Segmented line with its vertical projection showing the columns used for word segmentation

**Fig. 15**   Segmentation of a word (a) word image with horizontal projection of word taken to locate header line (b) word image after   removing header line with vertical projection to locate segmenting columns (c) segmented core characters with horizontal projection (d) final word after segmentation

**Fig. 16**   Example of devanagari characters  with varying height (a) Half characters with height less than *Avg_Ht*    (b) Character with height equal to *Avg_Ht*    (c) Height greater than *Avg_Ht*  with gap between core character and modifier (d) ) Characters with height greater than *Avg_Ht*  and joined core character and modifier

**Fig. 17**   (a) Standard vertical matra proportion before segmentation[24] (b) Vertical matra proportion before segmentation

**Fig. 18**   Segmentation of characters with joined lower modifier (a) Image $I_D$ of character consisting of lower modifier with dimensions HxW (b) character image with segmentation region within the rectangle (c) Difference value calculated for each row in segmentation region (d) Separated subimages of core character and lower modifier.

**Fig. 19** (a) Vertical projection of character image before removing lower modifier (b) Vertical projection of character image after removing lower modifier

**Fig. 20** Conjuncts and shadow characters: (a) Conjunct with combination of two characters (b) Conjunct with combination of three characters (c) Example of shadow characters (d) Image with characters in shadow and touching each other.

**Fig. 21** Analysing conjuncts on the basis of their width (a) characters with their respective width (b) conjuncts with 2 characters   (c) conjuncts with 3 characters

**Fig. 22** The image of conjunct: (a) Segmentation region in conjunct character (b) Image with resulting segmentation column

**Fig. 23** Flowchart for segmenting conjuncts.

**Fig. 24** Characters in shadow and their segmentation: (a) Characters in shadow (b) Image showing segmentation region within the rectangle (c) Segmentation column represented with dotted line (d) Individual characters after segmentation

**Fig. 25** (a) Example of shadow character with gap in constituent character shown with dotted circle (b) dilated image with number of core characters corresponding to number of CCs

**Fig. 26** Flowchart representing method to isolate shadow characters

**Fig. 27** (a) Dilated image with shadow character (b) constituents characters obtained after subtraction

**Fig. 28** Example of 8-connected neighbours with origin at *M(r,c)*

**Fig. 29** Top  strip components (a) with one CC (b) with two CCs

**Fig. 30** Modifier image showing extreme points

**Fig. 31** Example illustrating neighbouring features of two different  modifiers.

**Fig. 32** Grid showing 13 different non-uniform zones

**Fig. 33** (a) Character image partitioned in different zones (b),(c) mask corresponding to region